\documentclass[twoside,11pt]{article}

\usepackage{jmlr2e}
\usepackage[utf8]{inputenc}
\usepackage{placeins} 

\usepackage{dsfont}
\usepackage{natbib}
\usepackage{graphicx}
\usepackage{amsmath}
\usepackage{amssymb}
\usepackage{amsfonts}
\usepackage{mathdots}
\usepackage{xcolor}
\usepackage[plain,noend,ruled]{algorithm2e}
\usepackage{siunitx}
\usepackage{hyperref}

\DeclareMathOperator{\diag}{diag}

\DeclareMathOperator{\Cov}{Cov}

\newcommand{\tred}[1]{#1}

\newcommand{\dw}[1]{#1}

\ShortHeadings{Generalized Matrix Factorization}{Kidzi\'nski, Hui, Warton, and Hastie}
\firstpageno{1}

\begin{document}

\title{Generalized Matrix Factorization: efficient algorithms for~fitting generalized linear latent variable models\\to~large data arrays}


\author{\name \L ukasz Kidzi\'nski \email lukasz.kidzinski@stanford.edu \\
       \addr Department of Bioengineering\\
       Stanford University\\
       Stanford, CA 94305, USA
       \AND
       \name Francis K.C. Hui \email francis.hui@anu.edu.au\\
       \addr Research School of Finance, Actuarial Studies and Statistics\\
       The Australian National University\\
       Canberra, ACT 2601, Australia
       \AND
       \name David I. Warton \email david.warton@unsw.edu.au \\
       \addr School of Mathematics and Statistics\\
       and Evolution \& Ecology Research Centre\\
       The University of New South Wales\\
       Sydney, NSW 2052, Australia
       \AND
       \name Trevor Hastie \email hastie@stanford.edu \\
       \addr Department of Statistics and Biomedical Data Science\\
       Stanford University\\
       Stanford, CA 94305, USA
       }
\editor{}

\maketitle

\begin{abstract}%
Unmeasured or latent variables are often the cause of correlations between multivariate measurements, which are studied in a variety of fields such as psychology, ecology, and medicine. For Gaussian measurements, there are classical tools such as factor analysis or principal component analysis with a well-established theory and fast algorithms. Generalized Linear Latent Variable models (GLLVMs) generalize such factor models to non-Gaussian responses. However, current algorithms for estimating model parameters in GLLVMs require intensive computation and do not scale to large datasets with thousands of observational units or responses. 
In this article, we propose a new approach for fitting GLLVMs to high-dimensional datasets, based on approximating the model using penalized quasi-likelihood and then using a Newton method and Fisher scoring to learn the model parameters. \tred{Computationally, our method is noticeably faster and more stable, enabling GLLVM fits to much larger matrices than previously possible.
We apply our method on a dataset of 48,000 observational units with over 2,000 observed species in each unit and find that most of the variability can be explained with a handful of factors. We publish an easy-to-use implementation of our proposed fitting algorithm.}
\end{abstract}


\section{Introduction}\label{s:intro}
Latent factors underlying multivariate observations are of great interest in many applied disciplines. For example, in psychology or sociology, researchers measure multiple correlated test items to quantify certain constructs. They assume that responses can be described in terms of a small set of latent variables and that these latent variables can be interpreted as psychological or sociological traits \citep{bartholomew2011latent,skrondal2004generalized,spearman1904general}. In genetics, researchers measure gene expression in patients and expect that they correlate with intrinsic patient's features, often not measurable directly \citep{stegle2012using}. In ecology, researchers observe sets of species in independent observational units (e.g., sites) and assume the existence of latent features associated with the abundance of species
\citep[e.g., representing a low-dimensional community composition space][]{warton2015so,warton2016extending,ovaskainen2017make}.


Data from such experiments or observational studies can typically be expressed as a matrix of responses $Y = [y_{ij}]$, where rows $i \in \{1,2,...,n\}$ correspond to observational units (locations), subjects, etc. and columns $j \in \{1,2,...,m\}$ correspond to different responses such as species, genes, etc. Linear latent factors are then incorporated as a means of obtaining a low-rank approximation to the covariance structure of the rows in $Y$. \tred{Specifically, we assume that conditional on a set of latent variables $u_i \in \mathds{R}^p$, with $p \ll \min(m,n)$}, responses \tred{$y_{ij}$} are independent observations. See Section~\ref{s:model} for more detailed notation, and the model.

If all the responses are Gaussian in distribution, then we can find linear latent patterns with classical tools such as principal component analysis, singular value decomposition, or factor analytic models, and solutions can be computed very quickly even at scale. That is, many algorithms for fitting such large scale models have been suggested in the last two decades \citep{zou2006sparse,witten2009penalized,halko2011algorithm,hirose2015sparse,hirose2018sparse}. \tred{These algorithms are fast and scale well, since in this case there are known closed-form solutions which leverage from joint normality of the responses and latent variables.} However, such theory and computation for Gaussian responses does not generalize easily to non-Gaussian cases.

Generalized Linear Latent Variable Models \citep[GLLVMs,][]{moustaki2000generalized,niku2017generalized} are a class of models which generalizes factor analysis to non-Gaussian responses. Specifically, they assume that responses follow distributions in the exponential family, where the mean for each response varies as a function of the observed covariates and the aforementioned set of latent features via a known link function. Model parameters are then usually estimated using Bayesian or maximum likelihood methods. Among a wide variety of Bayesian tools, practitioners use general purpose modeling software such as \texttt{Stan} \citep{carpenter2017stan}, integrated nested Laplace approximations \citep[INLA,][]{rue2009approximate}, as well as Bayesian Markov Chain Monte Carlo estimation with more specific software designed for GLLVMs \citep{tikhonov2020joint,boral19}.

In this article, we focus on likelihood-based methods for GLLVMs. In particular, given the latent variable $u_i$ for an observational unit $i$, all the responses $y_{ij}$ are \tred{assumed to be conditionally independent of each other}. Thus, the likelihood can be expressed as a product of $m$ individual conditional likelihoods, after which we marginalize out the latent variables $u_i$ \tred{to obtain the marginal likelihood function}. The key problem in this approach is that the integral over $u_i$ does not have a closed-form \tred{for non-Gaussian responses} and must be estimated or approximated by some means. To solve this problem, a number of methods have been proposed, including the Laplace method or some variation thereof \citep{huber2004estimation,bianconcini2012estimation,niku2017generalized,robin2019}, numerical integration methods using adaptive quadrature \citep{rabe2002reliable}, and variational approximations \citep{hui2017variational,niku2019efficient}. 

While these approaches generally lead to very accurate solutions, they are computationally expensive for high-volume or high-dimensional problems and are difficult to parallelize, making them infeasible for solving
large-scale problems. An alternative is to look for  approximate solutions. \tred{One example of this comes from \citet{pichler2020new}, who proposed} \dw{estimating the multivariate joint likelihood using Monte Carlo sampling, along the lines of \citet{hui2015model}}, substantially improving computational performance thanks to the use of Graphical Processing Units (GPUs). To date their method has only been developed for binary responses. \dw{\citet{huber2004estimation} observed} that if the latent scores are treated as fixed parameters, then estimates of them obtained through the Laplace method coincide with variables derived using Penalized Quasi-Likelihood estimation (PQL)\dw{. This} is a well-establised method for estimating Generalized Linear Mixed Models \citep[GLMMs; see][for usages of PQL in the mixed model setting]{green1987penalized,breslow1993approximate,mcgilchrist1994estimation}\dw{, and it is of interest to look at how effective PQL might be in the GLLVM context}. 

In this article, we propose a new approximate solution to the problem of fitting GLLVMs, based upon existing ideas for estimating parameters in GLMMs. In particular, we develop an alternating method which
leverages the idea that a solution to a GLLVM can be approximated using Penalized Quasi-Likelihood estimation. This leads us to propose two algorithms: a direct Newton method with a simplified Hessian, and an Alternating Iteratively Reweighted Least Squares (AIRWLS) algorithm. As the name suggests, the AIRWLS algorithm uses classical Iteratively Reweighted Least Squares iterations by applying them alternately to rows and columns of $Y$. 

\tred{Our contributions are threefold. First, we design an algorithm that is orders of magnitude faster than existing methods for fitting GLLVMs, but with similar accuracy if responses matrices are sufficiently large. Moreover, our approach can be further sped up by distributing it across many machines. Second, thanks to the much-improved computational performance of our method, we can find latent-variable decompositions of large matrices, enabling new directions of research. This is illustrated on an ecological dataset on the coexistence of species across 48,000 observational units with over 2,000 responses in each unit. Third, we release easy-to-use \dw{software implementing the proposed approaches}, which can be used as a drop-in replacement for other linear latent variable modeling approaches.} 

\section{Model formulation} \label{s:model}
We begin by providing a more precise mathematical formulation of the GLLVM.
For observational unit $1\leq i\leq n$ and response $1 \leq j \leq m$, we assume
\begin{align}
y_{ij} | \mu_{ij} &\sim \mathcal{F}(\mu_{ij},\phi_j)\nonumber\\
g(\mu_{ij}) &= \eta_{ij} = \beta_{0j} + x_i^\intercal\beta_j + u_i^\intercal\lambda_j,\label{eq:model}
\end{align}
\tred{where $x_i \in \mathds{R}^d$ are a set of covariates observed for the $i$-th observational unit. The parameters $\beta_{0j}$ are intercepts for each response, and $\beta_j \in \mathds{R}^d$ are response-specific regression coefficients corresponding to the covariates. Next, the vectors $u_i\in \mathds{R}^p$ denote the latent variables (also referred to as factor scores) for observational unit $i$, with $\lambda_j \in \mathds{R}^p$ the corresponding response-specific factor loadings. We assume that distribution $\mathcal{F}(\mu_{ij},\phi_j)$ is from the exponential family of distributions with mean $\mu_{ij}$ and response-specific dispersion parameter $\phi_j$. The function $g(\cdot)$ is a known link function, e.g., the logit link for binary responses and log link for Poisson responses. Finally, we use $M^\intercal$ to denote the transposition of a matrix $M$.}

In the GLLVM formulated above, we assume that:
(A1) $u_i \sim \mathcal{N}(0,I_p)$ where $I_p$ is $p \times p$ identity matrix, 
(A2) $\Lambda = [\lambda_1,\lambda_2,...,\lambda_m]$ is a $p\times m$ lower triangular with positive elements on the diagonal, and 
(A3) all observational units are independent and, conditioned on $u_i$, the responses $y_{ij}$ are independent of each other. \tred{Critically, the second part of Assumption A3 implies that, conditional on $u_i$, the model \tred{as defined in~\eqref{eq:model}} specifies a Generalized Linear Model
\citep[GLM,][]{mccullagh1983glm} for each response~$j$ with mean $\mu_{ij}$ and dispersion parameter $\phi_j$.
\tred{Assumptions (A1) and (A2) are made in the context of GLLVMs to ensure parameter identifiability. In particular,} without constraining $\Lambda$ to be lower \dw{diagonal and} its diagonal elements to be positive, then} we could rotate the vectors $u_i$ and $\lambda_j$ without changing the value of the linear predictor $\eta_{ij}$ in \eqref{eq:model} \citep{huber2004estimation}. The independence of observational units as in Assumption (A3) is common in many formulations of GLLVMs \citep[e.g.,][]{hirose2015sparse,hui2017variational,niku2017generalized}.

Throughout the remainder of the article, we use matrix notation whenever it is convenient and suitable. In particular, apart from $\Lambda$ already defined, we use $B = [\beta_1,\beta_2,...,\beta_m], X^\intercal=[x_1,u_2,...,x_n]$ and $U^\intercal=[u_1,u_2,...,u_n]$ to define matrices corresponding to the regression coefficients, observed covariates, and factor loadings, respectively. We also denote the matrix of responses as $Y = [y_{ij}]$ and the matrix of means $M=[\mu_{ij}]$. \tred{Finally, letting $\Psi$ denote all the model parameters in the GLLVM, i.e. a vector formed from concatenating $\beta_{0,i}$, $\phi_j$, $\beta_j$, $\lambda_j$ for all $i$ and $j$, then we let $f(y_{ij}|u_i,\Psi)$ denote the probability density/mass} function of $y_{ij}$ given $u_i$ and $\Psi$, corresponding to the distribution $\mathcal{F}(\mu_{ij}, \phi_j)$ as defined \tred{in} \eqref{eq:model}. 

To conclude this section, we note that the usual factor analytic model can be expressed in terms of equation \eqref{eq:model} if we set $\mathcal{F}(\mu, \phi)$ to be a Gaussian distribution, and have $x_i$ represent only an intercept term. As the methods introduced in this paper for fitting \eqref{eq:model} in the general form rely on analogous matrix factorization concepts, then in turn, we will refer to them as \emph{generalized matrix factorization}.
  


\section{Likelihood estimation}
\tred{Let $f(y_{i}|u_i, \Psi)$ denote the conditional multivariate density function of the vector $y_i = (y_{i1},...,y_{im})^\intercal$ given the latent variables $u_i$. Then by the second part of Assumption (A3),} we have $f(y_{i}|u_i,\Psi) = \prod_j f(y_{ij}|u_i,\Psi)$. We then integrate out the random latent variables and write the marginal log-likelihood for the model defined in \eqref{eq:model} as:
\begin{align}
\ell(\Psi) &= \sum_{i=1}^n \ell_i(\Psi) = \sum_{i=1}^n \log\left( \int \prod_{j=1}^m f(y_{ij}|u_i,\Psi) \tred{\pi}(u_i)du_i \right),\label{eq:likelihood}
\end{align}
where $\tred{\pi}(u_i) \sim \mathcal{N}(0,I_p)$ and $\ell_i$ is the \tred{marginal log-likelihood of the $i$-th observational unit alone.}

Except for the special case where all the responses are Gaussian and the identity link function is used, the integral in \eqref{eq:likelihood} can not be expressed in closed form, and has thus led to an extensive amount of research into overcoming this computational burden. For example, it can be directly computed using numerical integration methods, including Gauss-Hermite quadrature, adaptive quadrature, or Monte-Carlo integration. In brief, quadrature approaches aim at approximating the integral as a $(2R-1)$ polynomial by evaluating the function at $R$ quadrature points. Adaptive quadrature shifts and scales locations of sampling points at each step to minimize the error \citep{rabe2002reliable}. Monte-Carlo methods sample the function within its domain and average the values. Importance sampling, a more sample-efficient method, was introduced for linear mixed models by \citet{kuk1999laplace} and \citet{skaug2002automatic} and can also be used here. While these methods can yield solutions close to exact, they are very expensive computationally and scale poorly with the number of latent variables.

Alternatively, we can approximate the value of the integral \eqref{eq:likelihood} using, say, a variational approximation or Taylor expansion (better known as the Laplace method), and maximize the approximated log-likelihood function instead. In the variational approximation, recently introduced in the context of GLLVMs by \citet{hui2017variational} and \citet{niku2019efficient}, Jensen's inequality is applied to obtain a lower bound for the log-likelihood that is (closed to) fully closed form and thus computationally manageable. Maximizing this variational lower bound then gives an approximate solution to \eqref{eq:likelihood}.
\tred{On the other hand, the Laplace method is based on taking a (typically second order) Taylor approximation of the log of the integrand in \eqref{eq:likelihood} around its mode, and then integrating the approximated integrand \citep{huber2004estimation,bianconcini2012estimation}.}
Although estimates based on using the Laplace method may exhibit a non-negligible finite sample bias relative to aforementioned methods such as Monte-Carlo methods, they are consistent to order $O(m^{-1})$ \citep{kass1995bayes,vonesh1996note}. 
%

\tred{While they are typically faster than their numerical integration counterparts, approximate likelihood methods such as the Laplace method still tend to be computationally prohibitive for problems where the response matrix consists of thousands of rows and/or columns, and may require days, weeks, or more time to converge (see \citealt{pichler2020new} and the empirical study in Section~\ref{ss:species} and Figure~\ref{fig:sim-results} for empirical demonstrations of this). Motivated by applications of GLLVMs and generalized matrix factorization to such high-dimensional settings, in this article, we leverage the structure of \eqref{eq:model} in two ways: (i) we show that by approximating the marginal log-likelihood using a Penalized Quasi-Likelihood approach, we can efficiently estimate all the required gradients and Hessians necessary for estimation of model parameters; (ii) thanks to this approximation, we demonstrate that the estimation problem can be decomposed to a set of smaller and relatively simple estimation problems based on individual rows and individual columns of the response matrix, and thus treated in an alternating fashion. As a result, our proposed is extremely parallelizable, thus enabling further performance optimization.}

\section{Penalized Quasi-Likelihood for GLLVMs} \label{s:pql}
In the pursuit of an efficient algorithm for estimating the model parameters in \eqref{eq:model}, we borrow ideas from \citet{huber2004estimation} who showed that, in the setting of GLLVMs, the maximum likelihood estimators of the latent variables derived from applying the Laplace method to the marginal likelihood in \eqref{eq:likelihood} are equivalent to those based on maximizing a Penalized Quasi-Likelihood \citep[PQL,][]{green1987penalized,breslow1993approximate}. On the other hand, the estimates of $\beta$ and $u_i$ are not equivalent and are indeed more biased for the PQL approach relative to the Laplace method. \tred{However, it has nevertheless been proven that PQL produces asymptotically consistent estimates of the model parameters and random effects provided the size of each observational unit, which in this case corresponds to the number of responses, grows with the number of units \citep{nie2007convergence,hui2017joint}.} This insight is particularly promising in the context of our work, since we focus on large-scale problems.

\tred{In what follows, we build on the derivation of PQL and Laplace methods proposed by \citet{breslow1993approximate} and \citet{huber2004estimation}. Specifically, in order to speed up computation by orders of magnitude, we propose to drop the slowly-varying terms arising from applying the Lapace method to the GLLVM marginal log-likelihood function, and use a diagonal approximation in place of Hessians. Furthermore, by alternating the optimisation between columns and rows, our methods become amenable to parallel computing. We show that the bias introduced by these modifications is small when the response matrices are sufficiently large. It is important to point out that, while PQL has been developed and studied to a large degree for GLMMs, it has never been formally established and applied to GLLVMs. To our knowledge, this article is the first to do so.}
  
\subsection{Derivation}\label{ss:pql-approx}
We start by considering the log-likelihood $\ell_i(\Psi)$ for the $i$-th observational unit. \tred{We follow standard GLM conventions popularized by \citet{mccullagh1983glm}. Furthermore, for ease of notation, we assume distributions in the exponential family with some known cumulant function $b(\cdot)$ and that the canonical link function is used.} The developments below can be extended to the case of a non-canonical link function, at the expense of additional tedious algebra. Let
\begin{align}
\exp(\ell_i(\Psi)) &= \int \prod_j^m h(y_{ij},\phi)\exp\left( \frac{y_{ij}\eta_{ij} - b(\eta_{ij})}{\phi_j}\right) \exp\left(-\frac{u_i^\intercal u_i}{2}\right) du_i\nonumber\\
&\propto \int  \exp\left( \sum_{j=1}^m \frac{y_{ij}\eta_{ij} - b(\eta_{ij})}{\phi_j}-\frac{u_i^\intercal u_i}{2}\right) du_i,\
\label{eq:quasi-likelihood-approx}
\end{align}
where $g(\mu_{ij}) = \eta_{ij}$ as defined in equation \eqref{eq:model}. We can express the integral \eqref{eq:quasi-likelihood-approx} in the form
$\int \exp(-L_i(u_i))du_i$ in order to apply the Laplace method. 
\tred{Specifically, let $\tilde{u}_i$ be the solution to $\partial L_i (u) / \partial u_i = 0$, i.e. the minimum of $L_i(u)$. Then the Laplace method yields}
\[
\ell_i(\Psi) \approx -\frac{1}{2}\log\left|\frac{\partial^2 L_i}{\partial u_i\partial u_i^\intercal}(\tilde{u}_i)\right| - L_i(\tilde{u}_i),
\]
where $|\cdot|$ denotes the determinant of a matrix.

\tred{In order to compute the derivatives of $L_i$ with respect to the latent variables $u_i$, we first calculate the partial derivatives of the summands in $L_i$. Let $v(\mu_{ij}) = 1/g'(\mu_{ij})$ be the variance function associated with the exponential family of distributions, 
e.g., for the Bernoulli distribution 
we have $v(\mu) = \mu(1-\mu)$. Then, assuming the canonical link function is used, we have}
\begin{eqnarray*}
  \frac{\partial \eta_{ij}}{\partial u_i}  = \frac{\mu^\intercal_{ij}(\eta_{ij})}{v(\mu_{ij})} = \lambda_j \text{ \ and \  }
  \frac{\partial b(\eta_{ij})}{\partial u_i}  = \frac{\mu^\intercal_{ij}(\eta_{ij})}{v(\mu_{ij})} = \lambda_j\mu_{ij}.
\end{eqnarray*}
Therefore, we obtain
\begin{align}
\frac{\partial L_i}{\partial u_i}(u_i) = -\sum_{j=1}^m \frac{\lambda_j( y_{ij} - \mu_{ij})}{\phi_j} + u_i\label{eq:kappa-prime}
\end{align}
and
\begin{align}
\frac{\partial^2 L_i}{\partial u_i\partial u_i^\intercal}(u_i) = \sum_{j=1}^m\frac{\lambda_j\lambda_j^\intercal v(\mu_{ij})}{\phi_j } + I_p. \label{eq:kappa-hessian}
\end{align}

Equation~\eqref{eq:kappa-hessian} can be rewritten as $\Lambda W \Lambda^\intercal + I_p$, where $\Lambda = [\lambda_1,...,\lambda_m]$ is the $p \times m$ matrix of factor loadings and $W$ is a
$m\times m$ diagonal matrix with elements $w_j~=~v(\mu_{ij}) / \phi_j$ for $j \in \{1,2,...,m\}$. Note that in the case of GLLVMs, the elements~$\{w_j\}_{j=1}^m$ correspond precisely to iterative weights coming from a GLM \citep[Chapter~2.5]{mccullagh1983glm}.

For the $i$-th observational unit then, we can write
\begin{align}
\ell_i(\Psi) \approx -\frac{1}{2}\log|\Lambda W\Lambda^\intercal + I_p| + \sum_{j=1}^m \frac{1}{\phi_j}(y_{ij}\tilde\eta_{ij} - b(\tilde\eta_{ij})) - \frac{1}{2}\tilde{u}_i^\intercal\tilde{u}_i\label{eq:ql-extended},
\end{align}
where $\tilde\eta_{ij} = \beta_{0j} + x_i'\beta_j + \tilde{u_i}'\lambda_j$. \tred{In the context of GLMMs and faced with a similar approximation to the log-likelihood of a cluster, \citet{breslow1993approximate} argued that since $W$ varied slowly as a function of the model parameters, then (for fixed $\Lambda$) the first term in \eqref{eq:ql-extended} could be ignored. In the setting of GLLVMs however, $\Lambda$ is not fixed. On the other hand, if we consider the normalized log-likelihood $\sum_{i}\ell_i/(mn)$, then we observe that for fixed $m$ this first term is asymptotically negligible as $n$ gets large. Moreover if both $m$ and $n$ are growing, then the first term in \eqref{eq:ql-extended} is asymptotically dominated by the second term; see also \citet{demidenko2013mixed} and \citet{hui2017joint}. This is also confirmed in our simulation study, reported in Section \ref{s:simulations}. Hence with applications of GLLVMs to high-dimensional datasets in mind, we chose also to (conveniently) ignore this term in our approximation so as to facilitate increases in computational efficiency. It is important to emphasize that our focus and thus methodological developments are driven by generalized matrix factorization for high-dimensional datasets where $n$ and/or $m$ are relatively large e.g., from the hundreds up to potentially tens of thousands. For small datasets, we recommend using more precise fitting methods such as numerical quadrature or (higher order) Laplace methods \citep{bianconcini2012estimation}.}

After omitting the log-determinant term, we thus conclude that for \tred{large samples} we can use the following approximation
\begin{align}
\log \int \prod_{j=1}^m f(y_{ij}|u_i,\Psi) \tred{\pi}(u_i)du_i \approx C + \sum_{j=1}^m \frac{1}{\phi_j}(y_{ij}\tilde\eta_{ij} - b(\tilde\eta_{ij})) - \frac{1}{2}\tilde{u}_i^\intercal\tilde{u}_i\label{eq:green},
\end{align}
where $\tilde{u}_i$ is a solution of \eqref{eq:kappa-prime} and $C$ is some constant as a function of the model parameters. Equation \eqref{eq:green} has precisely the form of a PQL when applied to latent variable models. Moreover, this result implies that, provided $m$ is sufficiently large, we can obtain a good approximation to the marginal log-likelihood function in \eqref{eq:likelihood} as
\begin{align}
L = -\sum_{i=1}^n L_i(\Psi) = -\sum_{i=1}^n \sum_{j=1}^m \frac{1}{\phi_j}(y_{ij}\tilde\eta_{ij} - b(\tilde\eta_{ij})) + \frac{1}{2} \sum_{i=1}^n u_i^\intercal u_i  \label{eq:pql}.
\end{align}


Importantly, we can solve the PQL as given by \eqref{eq:pql} very efficiently using a Newton algorithm as we demonstrate in the \dw{following} section. In particular, we discuss an approach inspired by iteratively reweighted least squares, where we alternate between the estimation of $U$ and $\Lambda$ (Section~\ref{ss:iterated-approx}). \tred{Then, we introduce heuristics for estimating Hessians, which allows for a direct Newton algorithm on all the parameters, and substantially reduces the computations in each iteration of the Newton algorithm (Section~\ref{ss:quasi-newton}).}

\subsection{Newton algorithms}\label{ss:newton-approx}
One approach to optimizing \eqref{eq:pql} is via alternating minimization \citep{robin2019} with respect to $U$ and $\Lambda$. For ease of notation, assume $\phi_j = 1$ is known e.g., in the case of Poisson and Bernoulli distributed responses. In our proposed iterative algorithms, estimates of dispersion parameters can be updated after each iteration of the Newton algorithm, if required. Specifically for estimating $\phi_j$ we can use a method of moments or maximum likelihood after each iteration of our proposed algorithms \citep{nelder1972generalized}.

\tred{ Assuming a canonical link and $\phi_j = 1$, equations \eqref{eq:pql} and \eqref{eq:kappa-prime} imply that the derivative of $L$ with respect to $u_i$ is given by}
\begin{eqnarray*}
\frac{\partial L}{\partial u_i } = \frac{\partial L_i}{\partial u_i } = -\sum_{j=1}^m (y_{ij} -\mu_{ij})\lambda_j + u_i.
\end{eqnarray*}


In the case of a non-canonical link a similar expression can be written involving additional weights.
Likewise,
\begin{eqnarray}\label{eq:hession}
\frac{\partial^2 L}{\partial u_i\partial u_i^\intercal } 
  &=& \sum_{j=1}^m v(\mu_{ij})\lambda_j\lambda_j^\intercal+I_p,
\end{eqnarray}
and noting that
$$\frac{\partial^2 L}{\partial u_i \partial u_k^\intercal } = 0\mbox{ for $k\neq i$}.
$$
Hence the Hessian has the block diagonal form
$d^2 L = \diag\left(H_1,H_2,\ldots,H_n\right)$ with $H_i~=~\partial^2 L / u_i\partial u_i^\intercal.$

\tred{Similarly, we can straightforwardly calculate gradients and Hessians of the PQL in \eqref{eq:pql} with respect to $\lambda_j$ and $\beta_j$ as follows}
\begin{eqnarray*}
  \frac{\partial L}{\partial \lambda_j } &=& -\sum_{i=1}^n (y_{ij} -\mu_{ij})u_i, \\
  \frac{\partial L}{\partial \beta_j } &=&-\sum_{i=1}^n (y_{ij} -\mu_{ij})x_i
\end{eqnarray*}
and
\begin{eqnarray*}
\frac{\partial^2 L}{\partial \lambda_j\partial \lambda_j^\intercal }
  &=& \sum_{i=1}^n v(\mu_{ij})u_iu_i^\intercal\\
\frac{\partial^2 L}{\partial \beta_j\partial \beta_j^\intercal }
  &=& \sum_{i=1}^n v(\mu_{ij})x_ix_i^\intercal\\
\frac{\partial^2 L}{\partial \beta_j\partial \lambda_j^\intercal }
  &=& \sum_{i=1}^n v(\mu_{ij})x_iu_i^\intercal.
\end{eqnarray*}
Furthermore,
$$\frac{\partial^2 L}{\partial \lambda_i \partial \lambda_k^\intercal } = \frac{\partial^2 L}{\partial \beta_i \partial \beta_k^\intercal }=\frac{\partial^2 L}{\partial \beta_i \partial \lambda_k^\intercal } = 0 \mbox{ for $k\neq i$}.$$
Note that for fixed $u_i$, \tred{we can view $(x_i^\intercal,u_i^\intercal)^\intercal$ as an enlarged covariate vector, with corresponding response-specific parameters $(\beta_j^\intercal,\lambda_j^\intercal)^\intercal$}. In turn, the update step for each parameter $\theta \in \{u_i,\lambda_j,\beta_j\}$ takes the form 
\begin{align}\label{eq:update_step}
\tred{\theta_{t+1} = \theta_{t} + s [-\nabla^2_{\theta} L(\theta_t)]^{-1}\nabla_\theta L(\theta_t),}
\end{align}
where $\theta_{t}$ is the estimator of $\theta$ in $t$-th iteration, $\nabla_\theta L$ is the gradient of $L$ with respect to $\theta$, and $\nabla^2_{\theta} L(\theta)$ is the corresponding Hessian of $L$ at $\theta$. \tred{The step size $s > 0$ \dw{should} be chosen carefully in order to guarantee convergence and speed up computation. In our implementations, we use Wolfe conditions \citep{wolfe1969convergence} to address this issue, although we acknowledge that other methods for selecting the step size might further improve performance of our methods.}

We now proceed to discuss two iterative algorithms for computing the update step in \eqref{eq:update_step} efficiently. The first approach uses Fisher scoring and produces an exact update, leveraging the fact that $\nabla^2_{\theta} L(\theta)$ can be approximated (in fact, it is exact in the case where a canonical link is used) by the Fisher information matrix plus an identity matrix. The second approach uses only the diagonal of the Hessian $\nabla^2_{\theta} L(\theta)$, which is an approximate update that can be computed very quickly. In both approaches, after each update, we rotate the matrices $U$ and $\Lambda$ so as to satisfy the identifiability assumptions in Assumptions (A1) and (A2).

\subsubsection{Fisher scoring and Alternating Iteratively Reweighted Least Squares}\label{ss:iterated-approx}\label{ss:airwls}

One approach for optimizing the PQL is to leverage the fact that, when we use the canonical link function, the negative Hessian matrices as defined in Section~\ref{ss:newton-approx} are equivalent to the Fisher information \citep{nelder1972generalized} derived from the PQL in \eqref{eq:pql}. 
\tred{Specifically, conditional on $(\beta_j,\lambda_j)$ for all $j = 1,\ldots,m$, we can update each of the latent variables $u_i,\;i=1,\ldots,n$ in \eqref{eq:pql} by solving $n$ separate penalized GLMs \citep{green1987penalized,breslow1993approximate}, where in each GLM the vector of $m$ responses $(y_{i1}, \ldots, y_{im})^\top$ i.e., rows of $Y$, are treated as the ``observations''. Conversely, conditional on $u_i$ for $i = 1,\ldots,n$, we can update the response-specific coefficients and loadings $(\beta_j,\lambda_j)$ for $j=1,\ldots,m$ by solving $m$ separate (unpenalized) GLMs, where the model matrix for each GLM is formed from the expanded predictor vectors $(x_i^\intercal,u_i^\intercal)^\intercal$ and the responses are given by the columns of $Y$. }

Maximum likelihood for a single GLM is typically performed using the Newton algorithm implemented via iteratively reweighted least squares (IRWLS), or penalized IRWLS if a penalty is involved. Hence, minimizing the PQL in \eqref{eq:pql} can be accomplished by applying alternating, parallel IRWLS algorithms applied to the rows and columns of the response matrix.

We now illustrate the derivation of the $t$-th IRWLS update step for $u_i$ in more detail. Let $u_i^{(t)}$ be the estimate of $u_i$ in the $t$-th iteration, $I(u_i^{(t)})$ be the (penalized) Fisher information matrix, and $W_t$ be the iterative weight $m$-vector in the $t$-th iteration as defined above \eqref{eq:ql-extended}. Let $\mu_i^{(t)} = (\mu_{i1},\ldots,\mu_{im})^\intercal$ be the $i$-th row of the matrix of means estimated in the $t$-th iteration, and similarly let $Y_i = (y_{i1},\ldots,y_{im})^\intercal$ be the $i$-th row of the response matrix. We first derive the update step~\eqref{eq:update_step} with step size $s=1$.
\begin{align*}
  u_i^{(t+1)} &= u_i^{(t)} + [I(u_i^{(t)})]^{-1}\nabla L(u_i^{(t)}),\\
  &= u_i^{(t)} +  [\Lambda W_t \Lambda^\intercal+I_p]^{-1}[\Lambda(Y_i - \mu_i^{(t)}) - u_i^{(t)}],\\
  &= [\Lambda W_t \Lambda^\intercal+I_p]^{-1}\Lambda W_t[ \Lambda^\intercal u_i^{(t)} + W_t^{-1}(Y_i - \mu_i^{(t)})] \\
  &= [\Lambda W_t \Lambda^\intercal+I_p]^{-1}\Lambda W_tZ_t,
\end{align*}
where  $Z_t = \Lambda^\intercal u_i^{(t)} +  W_t^{-1}(Y_i - \mu_i^{(t)})$ is a working response vector. Hence we obtain $u_i^{(t+1)}$ by a ridge regression of $Z_t$ on $\Lambda^\intercal$ with weights $W_t$. When the step size $s\neq 1$, our update is instead $u_i^{(t)}+s(u_i^{(t+1)}-u_i^{(t)})$. Note that although the $\beta_j$'s (which are assumed to be fixed in the update) do not appear explicitly in these equations, the $j$th element of $\mu_i^{(t)}$ includes $x_i^\intercal\beta_j$ as part of its linear predictor i.e., an \emph{offset} in GLM parlance.

We use a similar procedure for updating $\lambda_j$ and $\beta_j$, \tred{conditional on the $u_i$'s. Specifically, the problem of estimating the response-specific coefficients and loadings} in
\[
g(\mu_{\cdot,j}) = X \beta_j + U \lambda_j,
\]
where $\mu_{\cdot,j} = (\mu_{1j},\ldots,\mu_{nj})^\intercal$ and $U^\intercal = [u_1,...,u_n]$ can be rewritten as
\[
g(\mu_{\cdot,j}) = (X, U) \gamma,
\]
where $(\cdot,\cdot)$ stands for horizontal concatenation of matrices and $\gamma = \left(\beta_j^\intercal, \lambda_j^\intercal\right)^\intercal$. Again we can solve it using IRWLS, although this time without the penalty term and with $(X,U)$ in place of $\Lambda$. At the end of each iteration, we rotate $U$ and $\Lambda$ to fulfill the identifiability requirements given as part of Assumptions (A1)-(A2). For the full summary of our alternating two-step procedure, we refer to Algorithm~\ref{alg:airwls}. Importantly, note that for each $i \in \{1,...,n\}$ in the PQL in \eqref{eq:pql}, the optimization problem can be decoupled and solved independently, \tred{allowing for the use of parallel computation to update the latent variables. Similarly, for each $j \in \{1,...,m\}$ in \eqref{eq:pql}, the optimization problem can be decoupled and the updates of the response-specific coefficients and loadings can be performed in parallel.}

\begin{algorithm}\label{alg:airwls}
\caption{\textsc{Alternating Iteratively Reweighted Least Squares}}
\begin{enumerate}
    \item Initialize $U,B,\Lambda$ randomly, where $B = [\beta_1,\beta_2,...,\beta_m]$.
    \item Repeat until the convergence condition:
    \begin{enumerate}
        \item Perform one step of \tred{penalized} IRWLS to regress rows of $Y$ on $\Lambda$. Store regression coefficients as~$U$.
        \item Perform one step of \tred{(unpenalized)} IRWLS to regress columns of $Y$ on $(X,U)$. Store regression parameters as $(B^\intercal,\Lambda^\intercal)^\intercal$.
        \item Transform data to comply with Assumptions (A1)-(A2):
          \begin{enumerate}
            \item Find a rotation $\Theta$ such that $\Cov(U\Theta) = I_p$, using, for example, principal component analysis,
            \item Compute $U_0 = U\Theta$ and $\Lambda_0 = \Theta^{-1}\Lambda$,
            \item Find a QR decomposition of $\Lambda_0^\intercal = QR$\tred{, such that $R$ has positive elements on the diagonal},
            \item Return $R^\intercal$ and $U_0Q$ as new estimates of $\Lambda$ and $U$ respectively.
          \end{enumerate}

    \end{enumerate}
\end{enumerate}
\end{algorithm}
\tred{As the convergence criterion} in Algorithm \ref{alg:airwls}, in our implementation we use the change in log-likelihood relative to the new log-likelihood, i.e., we stop when $|L_{k-1}-L_{k}|/|L_k| < \varepsilon$, where $L_k$ is the log-likelihood in $k$-th iteration and $\varepsilon$ is a sufficiently small value.

\subsubsection{Quasi-Newton with diagonal Hessians}\label{ss:quasi-newton}

Instead of using the Fisher information, we can, of course, derive the updated step directly from \eqref{eq:update_step} by computing and inverting the Hessian explicitly. For each of the  parameters $u, \lambda,$ and $\beta$, the corresponding Hessian matrices are block diagonal in structure. On the other hand, computing a Newton step requires inverting all blocks, which would be computationally expensive for large $n$ since the Hessians are $np\times np$ or $mp\times mp$ matrices. To speed up computation then, we propose an alternative quasi-Newton method where we only use the diagonals of blocks in Hessians. These diagonals and their inverses in \eqref{eq:update_step} can be computed \tred{quickly, and indeed our} empirical study later on shows that this approximation reduces computation time, despite increasing the number of steps required for convergence.

Note that the diagonal elements of \eqref{eq:hession} can be computed by taking
\begin{eqnarray}\label{eq:newton_gradients}
  \diag\left(\frac{\partial^2 L}{\partial u_i\partial u_i^\intercal }\right)
  &=& \diag\left(\sum_{j=1}^m v(\mu_{ij})\lambda_j\lambda_j^\intercal+I_p\right)\nonumber\\
  &=&  (\Lambda \circ \Lambda) v(\mu_{i,\cdot})^\intercal + \mathds{1}_p,
\end{eqnarray}
where $\circ$ denotes the Hadamard product, $\diag(\cdot)$ denotes the diagonal of the given matrix, $v(\mu_{i,\cdot}) = (v(\mu_{i1}), \ldots, v(\mu_{im}))^\intercal$, and $\mathds{1}_p = [1,1,...,1]^\intercal$. The diagonal elements of Hessians of $\lambda_j$ and $\beta_j$ are derived analogously. Specifically, we obtain
\begin{eqnarray}\label{eq:newton_hessians}
  \diag\left(\frac{\partial^2 L}{\partial \beta_j\partial \beta_j^\intercal }\right)
  &=&  (X^\intercal \circ X^\intercal) v(\mu_{\cdot,j}),\nonumber\\
  \diag\left(\frac{\partial^2 L}{\partial \lambda_j\partial \lambda_i^\intercal }\right)
  &=&  (U^\intercal \circ U^\intercal) v(\mu_{\cdot,j}),
\end{eqnarray}
where $v(\mu_{\cdot,j}) = (v(\mu_{1j}), \ldots, v(\mu_{nj}))^\intercal$. The full quasi-Newton approach thus follows the same steps as in Algorithm \ref{alg:airwls}, except for steps 2(a) and 2(b) where we replace the AIRWLS update with an explicit implementation of \eqref{eq:update_step} with Hessians given by equations \eqref{eq:newton_gradients}  and \eqref{eq:newton_hessians}.

\subsection{Regularized generalized matrix factorization}\label{ss:reg-gmf}




The above fitting algorithms for estimating GLLVMs and performing generalized matrix factorization has assumed that the number of latent variables, $p$, is known. In practice, however, we do not know the dimension of the latent space and we may need to estimate it from the data also. Methods for selecting the dimension \dw{include:} cross-validation\dw{;} information criteria \citep{bai2002determining,hirose2015sparse}\dw{;} a somewhat arbitrary choice for the threshold of the variance explained \citep{smith2015factor}\dw{;} or \dw{a} sparsity-inducing penalty added to the log-likelihood \citep{hui2018order}. We propose another smooth shrinkage parameter, motivated by regularized matrix factorization \citep{zou2006sparse}.

\begin{figure}[ht!]
    \centering
    \includegraphics[width=0.49\linewidth]{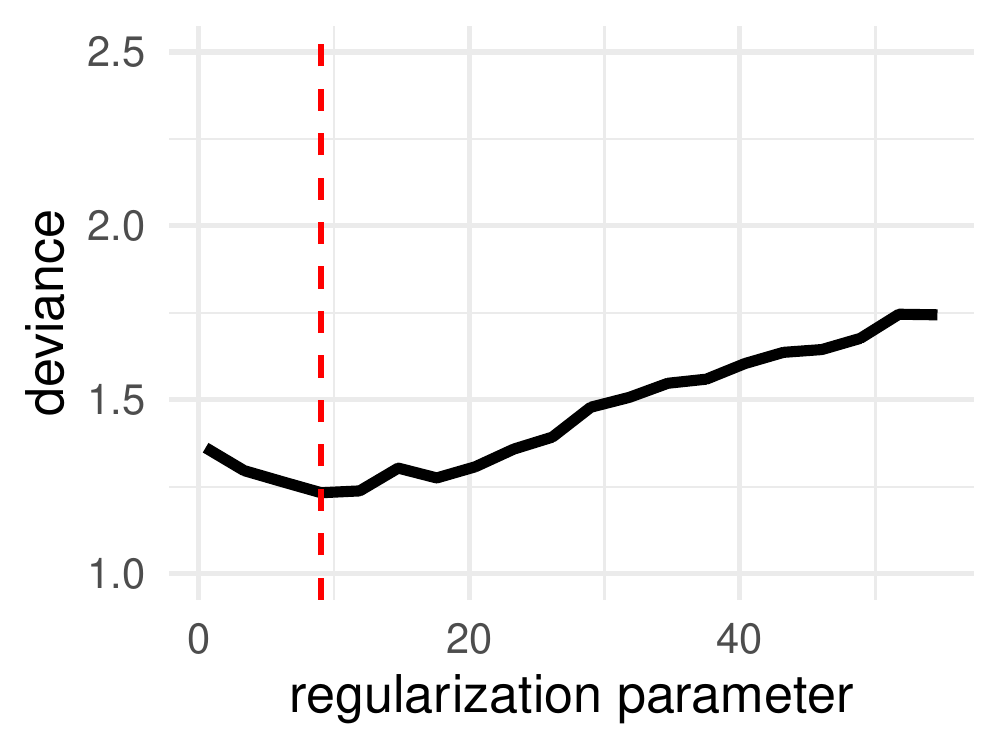}
    \includegraphics[width=0.49\linewidth]{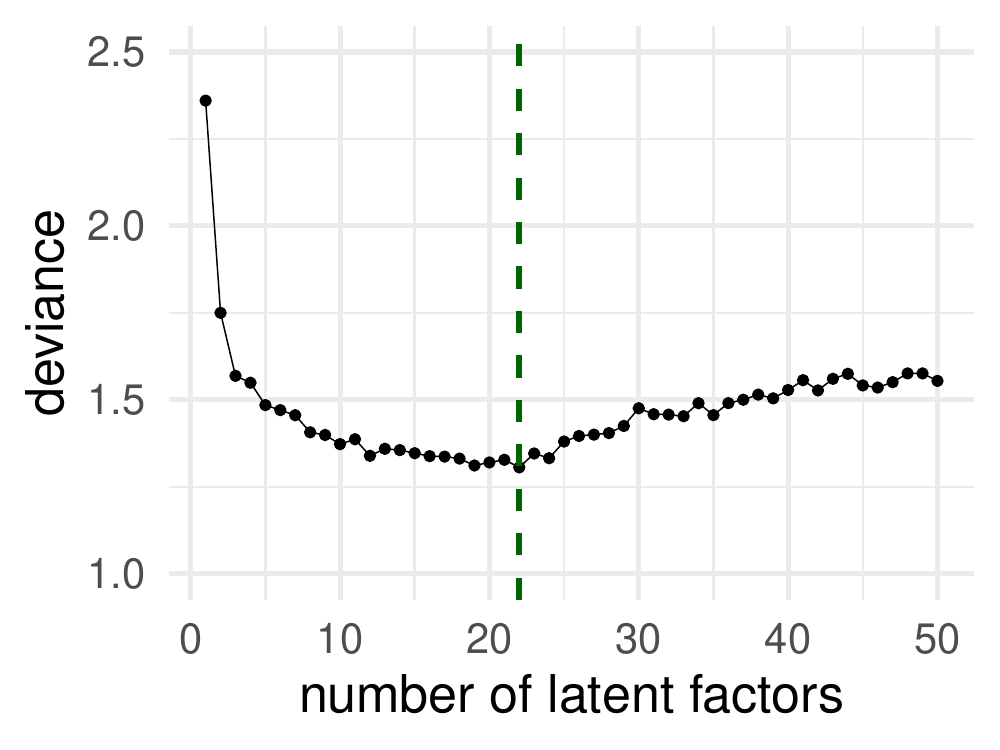}\\
    \includegraphics[width=0.49\linewidth]{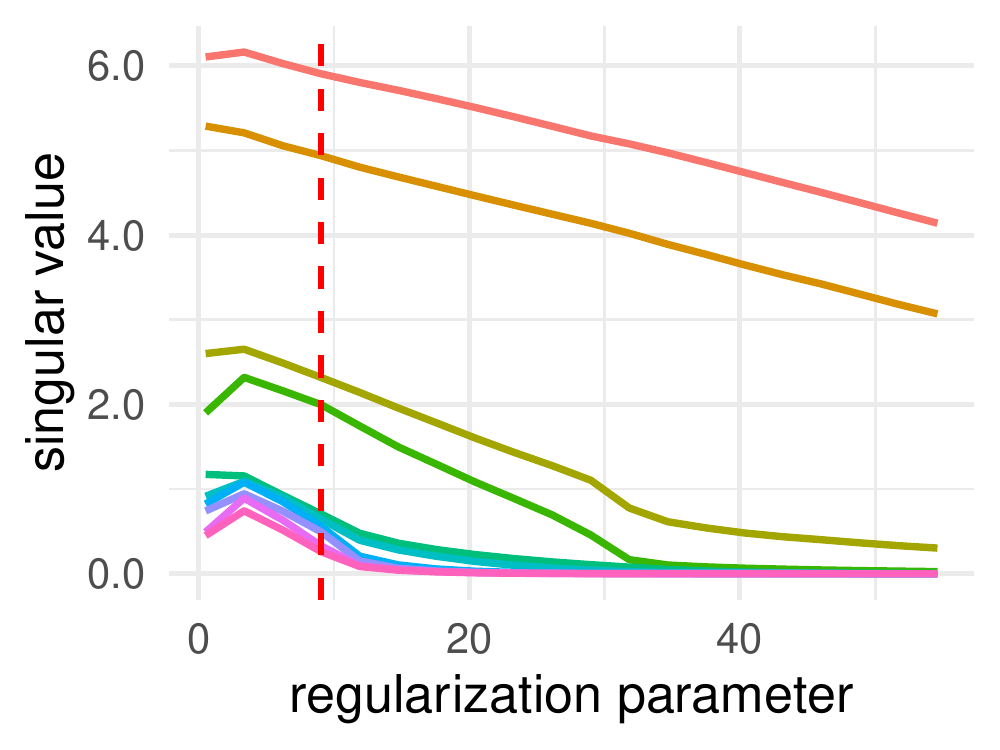}
    \includegraphics[width=0.49\linewidth]{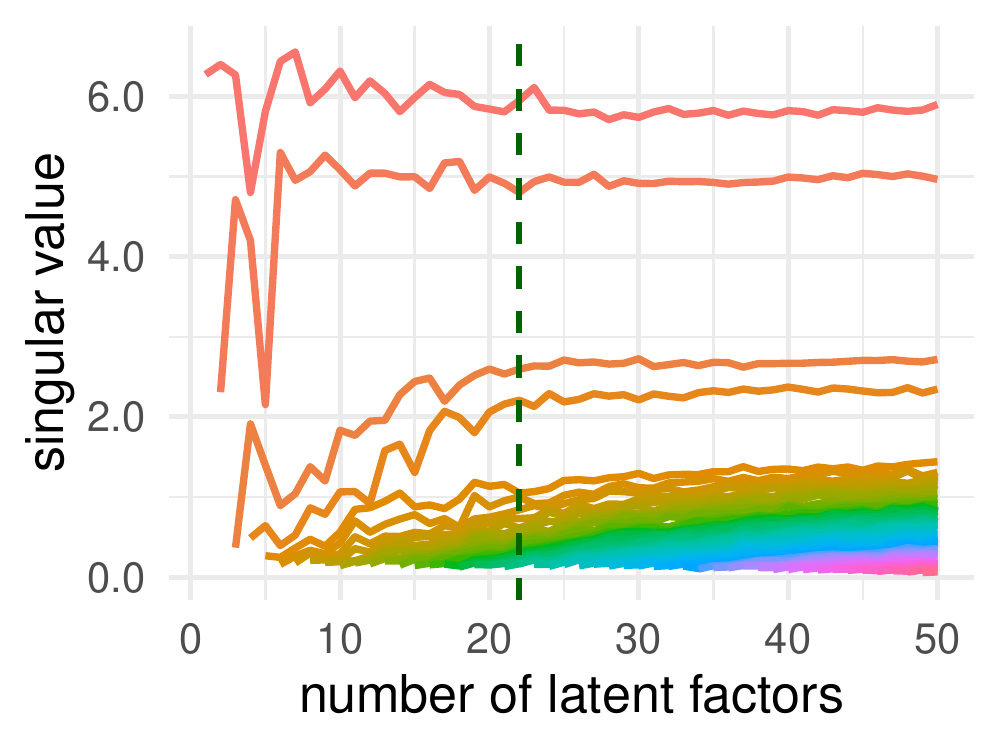}
    \caption{Model selection using the cross-validation framework. We present two methods for choosing the complexity of the model: smooth regularization (left panels) and rank constraint (right panels). \tred{The 20-fold cross-validation allowed us to select hyper-parameters minimizing the deviance (dashed lines).} The number of nonzero singular values for a given $\gamma$ is the selected rank. As expected\dw{, in \tred{the smooth regularization model} (bottom left), the singular values all shrank} when the regularizing parameter increased, while they \dw{remained stable when we added} more factors in \tred{the rank-constrained model} (bottom right).}\label{fig:latent-space-lasso}\label{fig:reg-gmf}
\end{figure}

Instead of controlling the rank by explicitly choosing the number of
latent variables, we can set a large upper bound on the number of
latent variables (e.g.~$\sqrt{m}$) and then  regularize the  latent
variables with an extra term $\frac{1}{2}\|\Lambda\|_2^2$ added to the
PQL criterion. If we control penalties with a scaling parameter $\gamma$, then this leads to the regularized objective function
\begin{align}
L_2(\Psi) = \sum_{i=1}^n\sum_{j=1}^m (y_{ij}\hat\eta_{ij} - b(\hat\eta_{ij})) + \frac{\gamma}{2}\|U\|_2^2 + \frac{\gamma}{2}\|\Lambda\|_2^2,\label{eq:two-penalties}
\end{align}
where $\|\cdot\|_2$ is the Frobenius norm. \tred{Note that since the variance of $u_i$ was arbitrarily set to $I$ in (A1), the two penalties on the norms of $U, V$ can be controlled with a single parameter $\gamma$ without loss of generality.}
\citet{srebro2005generalization} show that solving \eqref{eq:two-penalties} with $U$ and $\Lambda$ of sufficiently high rank is equivalent to solving
\begin{align}
L_*(\Psi) = \sum_{i=1}^n\sum_{j=1}^m (y_{ij}\hat\eta_{ij} - b(\hat\eta_{ij})) + \gamma\|M\|_*,\label{eq:nuclear}
\end{align}
where $\|\cdot\|_*$ denotes the  nuclear norm and $M = U\Lambda^\intercal$. Equation \eqref{eq:nuclear} can be interpreted as a relaxed version of a rank constraint on $M$ \citep[Section 7.3.4,][]{hastie2019statistical}. This relaxation makes the problem convex. To tune dimensionality, we can control the penalty parameter $\gamma$ in \eqref{eq:two-penalties}. In particular, for sufficiently large $\gamma$ some singular values of $M$ vanish, effectively reducing dimensionality of the latent space (See Figure~\ref{fig:latent-space-lasso} for an illustrative example).

We illustrate empirical properties of this method of regularized generalized matrix factorization as part of our simulation study in Section~\ref{ss:unknown-d}.


\section{Evaluation}\label{ss:evaluation}\label{ss:missing-data}
A baseline method for comparing the proposed algorithms for fitting GLLVMs is the \texttt{R} package \verb|gllvm| \citep{niku2019gllvm}, applied with its default settings.  This is a state-of-the-art package \tred{which uses variational approximations \citep{hui2017variational} in conjunction with automatic differentiation \citep{niku2019efficient} to perform maximum approximate likelihood estimation for GLLVMs in a computationally efficiency manner.} Since other approaches to the estimation of parameters in GLLVM are comparable or worse in terms of speed, then we only use the \verb|gllvm| package for comparison in our numerical studies below.

We use a series of metrics and techniques to compare the quality of the GLLVM fits using the estimation methods proposed in this article, relative to fits obtained using the \texttt{gllvm} package. This is important because although the above algorithms minimize a version of penalized deviance, in certain applications, practitioners might instead be interested in estimates of the fixed effect parameters, the accuracy of the generalized matrix factorization in constructing the latent space, or out-of-sample predictive performance of the model. 

\paragraph{Deviance for evaluating fit to responses.}
Following \citet{nelder1972generalized}, we define \tred{deviance of the GLLVM as
\[
D(Y,\hat M) = 2\sum_{i=1}^n\sum_{j=1}^m \left( \log f(y_{ij} | \hat{\mu}_{ij}, \hat{\phi}_j) - \log f(y_{ij} | \hat{\mu}_{0,ij}, \hat{\phi}_{0,j})\right),
\]
where the $\hat{\mu}_{ij}$'s and $\hat{\phi}_j$'s are the predicted values and estimated dispersion parameters from the fitted GLLVM, the $\hat{\mu}_{0,ij}$'s and $\hat{\phi}_{0,j}$'s are the predicted values and dispersion parameters based on the saturated GLLVM, and $f(y_{ij} | \mu_{ij}, \phi_j) = f(y_{ij} | u_i, \Psi)$ is the probability density/mass function of the responses assumed in \eqref{eq:model}.}

The absolute value of deviance is usually hard to interpret. Therefore, we choose to calculate the ratio of deviances between either two fitted models or between the fitted model and the null model. \tred{For a null model denoted as $M_{null}$, we refer to $1 - D(Y,\hat{M})/D(Y,M_{null})$ as a fraction of null deviance explained by $\hat{M}$. The closer the fraction is to one, the more deviance is explained by the fitted GLLVM.}

\paragraph{Procrustes error for evaluating fit of the latent space.} In certain applications we are interested in how accurate the generalized matrix factorization method recovers the latent space. Since vectors spanning latent spaces are not unique, then we use a metric that rotates them before comparison. In particular, we follow \citet{niku2019efficient} and (for the purpose of this article) define the Procrustes error as
\[
P(\Lambda_0, \hat\Lambda) = \min_{\Omega}\|\Lambda_0 - \Omega \hat\Lambda\|_F, \text{ subject to } \Omega^\intercal\Omega = I,
\]
where $\|\cdot\|_F$ is the Frobenius norm, $\Lambda_0$ denotes for the true value of the loading matrix, $\hat\Lambda$ is the estimated loadings, and $\Omega$ is a rotation matrix. Note that unlike the deviance metric defined above, the Procrustes error is only computable in simulation studies where we have access to the ground-truth factor loadings. 

\paragraph{Mean squared errors for evaluating fixed effect coefficients.} \tred{As a basic measure of the accuracy in estimating the fixed effects in the GLLVM, i.e., the $\beta_j$'s in \eqref{eq:model}, we compute the mean squered \dw{error} of estimated values with respect to true values as a normalized Frobenious norm,
\[
MSE(B_0,\hat B) = \|B_0 - \hat B\|_2^2 / md,
\]
where $d \times m$ matrix $B_0$ denotes the true value of the regression coefficient matrix, and $\hat{B}$ denotes the estimated value. As with Procrustes error, this metric requires access to the true $B_0$ and, as such, it is also only applicable in simulation studies. }

\paragraph{Predictive performance}
In certain situations, \tred{some elements of the response matrix $Y$ may not be observed and need to be predicted. Such predictions are relatively straightforward using our proposed algorithms and involve \dw{little additional} computational expense. For example, the AIRWLS algorithm} introduced in Section~\ref{ss:iterated-approx} can be used in missing data settings since, in each regression step, we can use only a subset of rows or columns as long as there are enough observations. That is, it can be used for sparsely observed multi-response data. For the quasi-Newton method proposed in Section~\ref{ss:newton-approx}, in order to compute gradients we need the fully observed response matrix. However, following ideas from the \textsc{Soft-Impute} method \citep{mazumder2010spectral}, in each iteration we can use predictions from the previous step to \emph{impute} missing values and then compute gradients. 

The above feature of our proposed algorithms also enables us to straightforwardly employ cross-validation for assessing the fit and choosing model tuning parameters e.g., tuning $\gamma$ in the regularized generalized matrix factorization in Section \ref{ss:reg-gmf}. For instance, to assess overall out-of-sample goodness-of-fit we can randomly sample elements of the observed response matrix $Y$, hold them out, and compute the out-of sample
deviance of the predictions i.e., computing the measure $D(Y,\hat M)$ defined above. Depending on the objective, multiple techniques for sampling can be used, including uniform sampling of matrix entries, sampling based on response values, or sampling based on values of predictors. We use this method both for evaluating and comparing GLLVM fits on real datasets (Section~\ref{ss:species}), and in simulations for choosing the optimal shrinkage parameter (Section~\ref{ss:unknown-d}).

\paragraph{Uncertainty.} \tred{Resampling and simulation methods can be used both to assess uncertainty of the estimated parameters, as well to as quantify prediction uncertainty. For instance, we can repeatedly simulate data from a fitted model (i.e., a form of parametric bootstrap, fit models to the simulated data, and use the empirical covariance matrix based on the bootstrapped model parameter estimates as an estimate of uncertainty. Inference such as Wald-based confidence intervals or hypothesis tests for one or more model parameters then follows from this. For measuring variability of predictive performance, we can randomly sample elements of the observed response matrix, hold them out, and compute deviance on the predictions. Alternatively, instead of randomly sampling elements, we can hold out entire rows of the response matrix and compare predictions based only on the fixed effects \citep{warton2008penalized,wang2012mvabund}. Finally, we can sample rows or columns with replacement, to estimate uncertainty of estimates of the fixed effects. Importantly, regardless of the precise resampling or simulation technique employed, such an approach to quantifying uncertainty is computationally feasible thanks to the substantially reduced computation time of the proposed approach to fitting GLLVMs, relative to classical methods.}




\section{Data studies}\label{s:data}

We report two applications from Ecology, with two distributions of responses: Poisson (Section~\ref{ss:ants}) and Bernoulli (Section~\ref{ss:species}). \tred{In the first study we analyze a $41\times30$ matrix of ant species, while in the second one we analyze a much larger \num[group-separator={,}]{48331} $\times$ 2,211 matrix of plant species}. We compare three methods for estimating GLLVMs: AIRWLS as described in Section~\ref{ss:iterated-approx}, the quasi-Newton method as discussed in Section~\ref{ss:quasi-newton} and which we refer to it as \emph{Newton}, and the variational approximations approach using the \texttt{gllvm} package \citep{niku2019gllvm}.

\subsection{Study 1: Abundance of ants}\label{ss:ants}

\begin{figure}[ht!]
    \centering
    \includegraphics[width=0.49\linewidth]{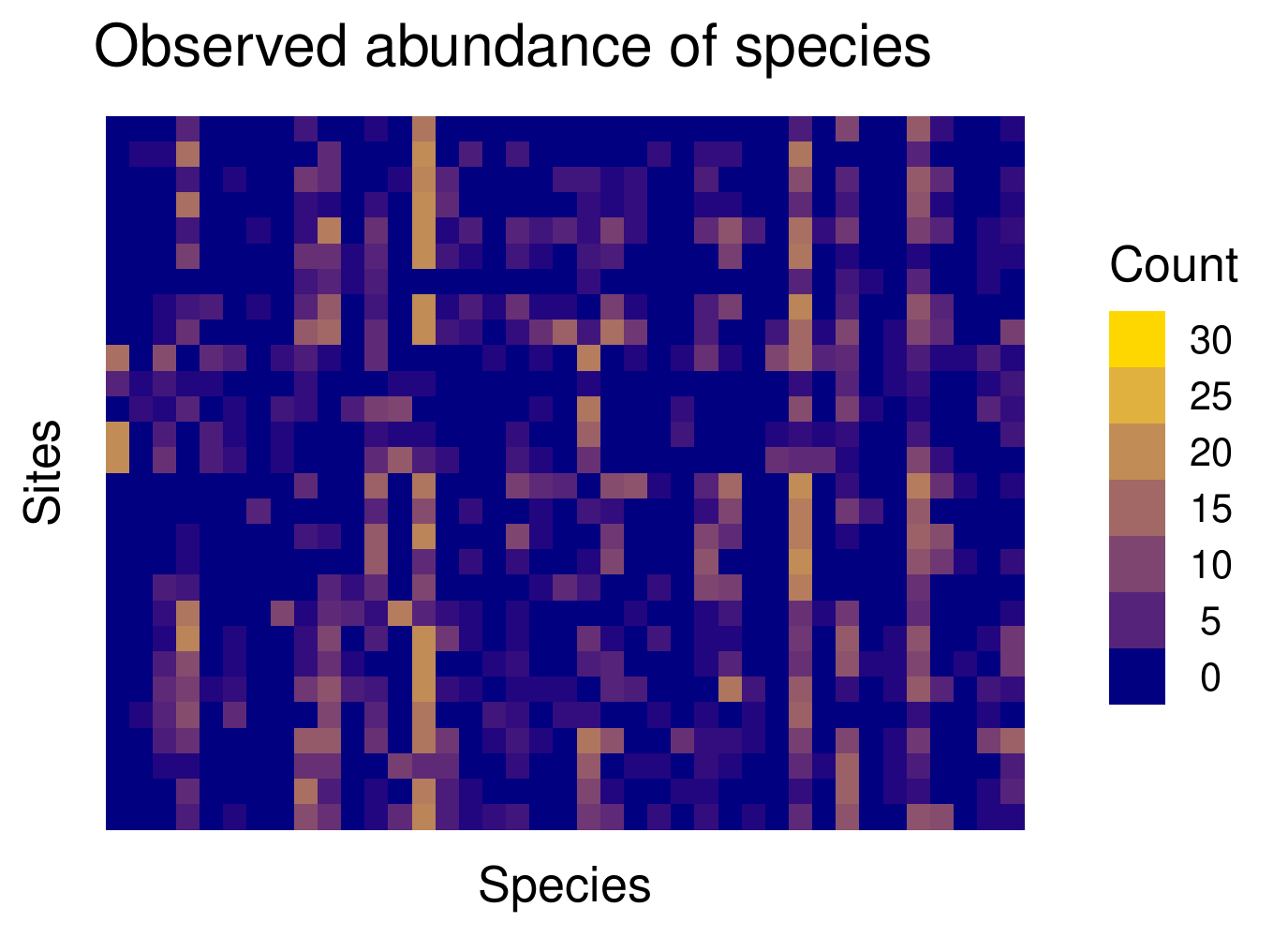}
    \includegraphics[width=0.49\linewidth]{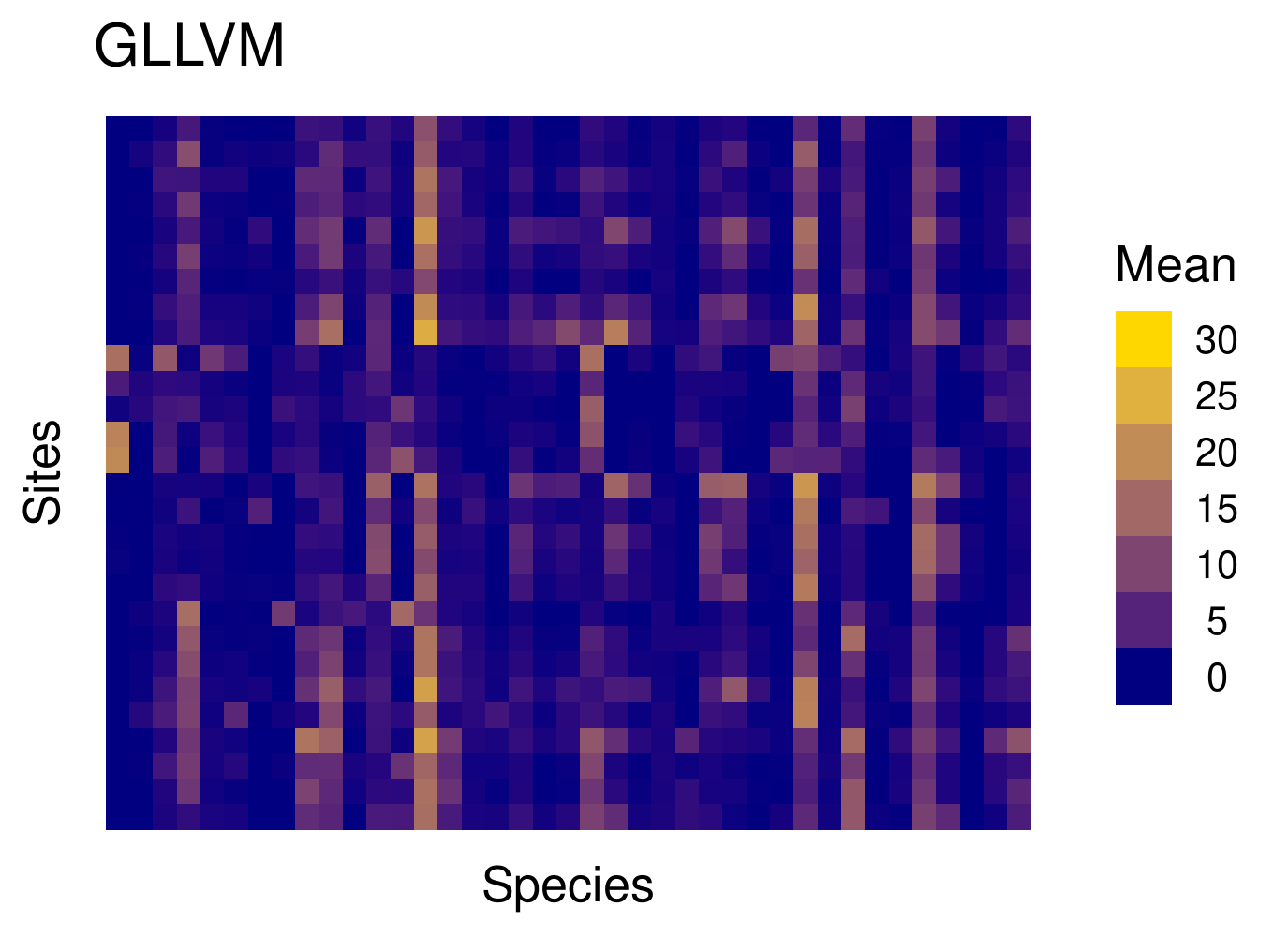}\\
    \includegraphics[width=0.49\linewidth]{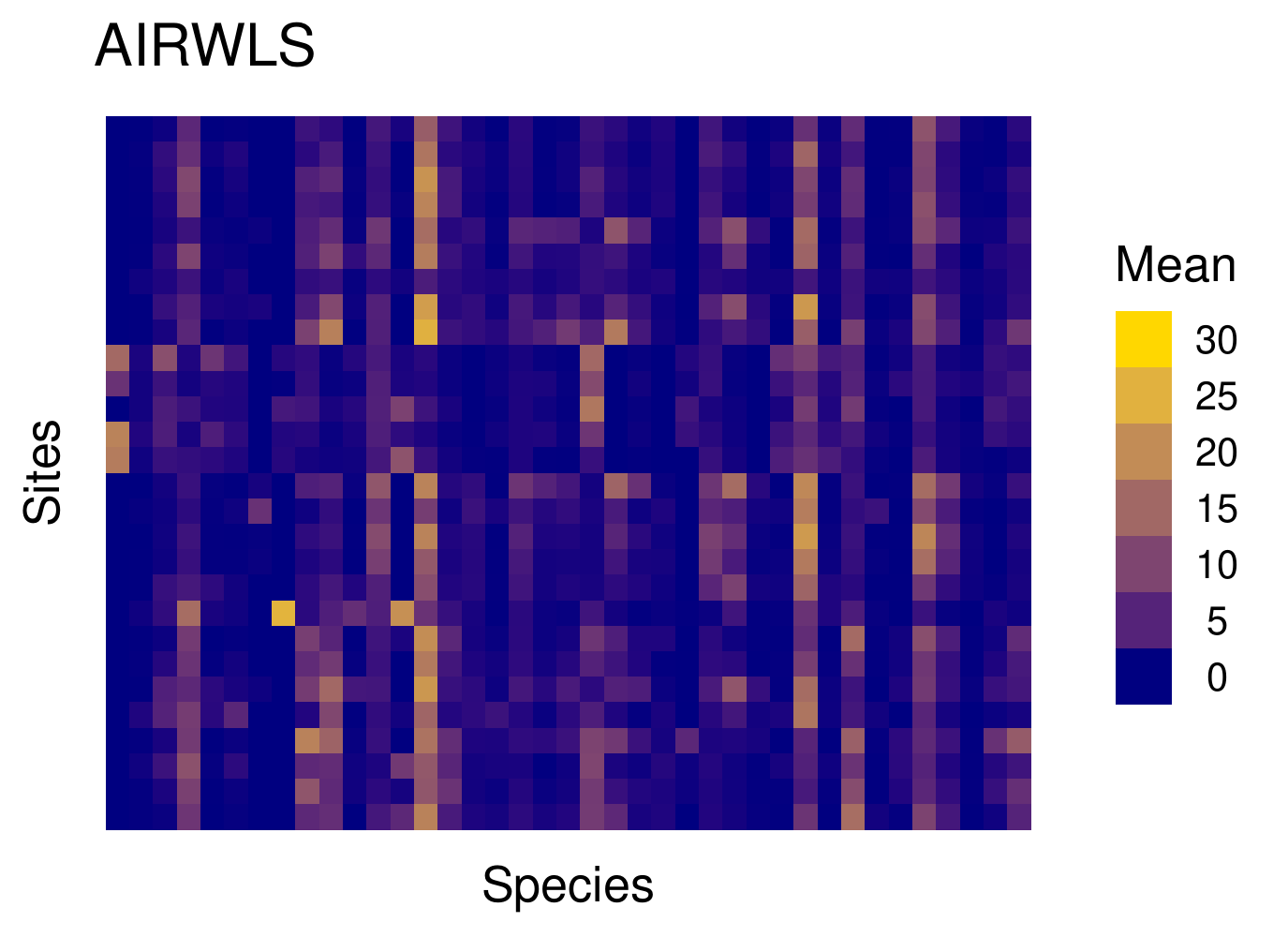}
    \includegraphics[width=0.49\linewidth]{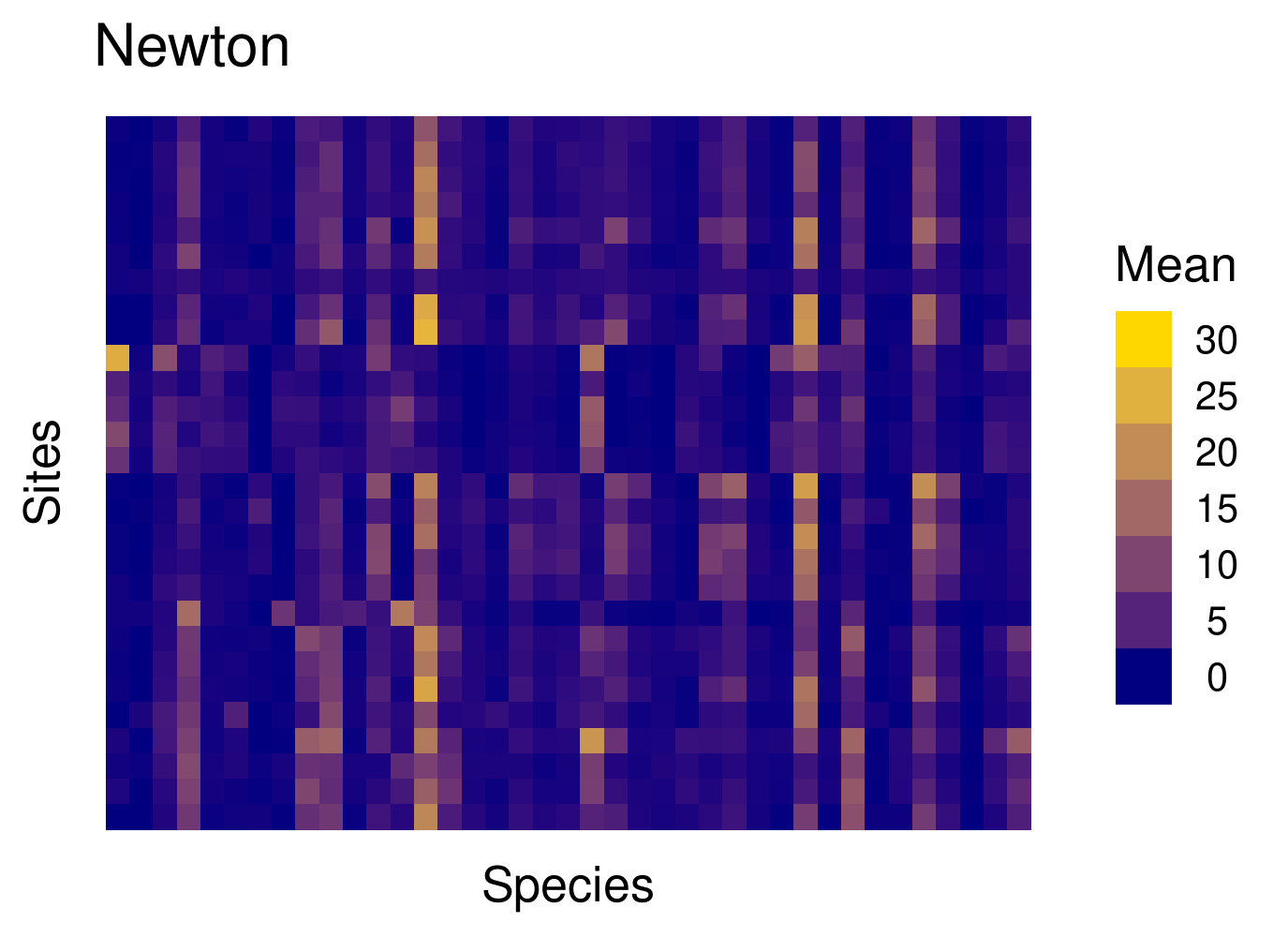}
    \caption{Qualitative validation of our algorithms on a dataset of abundance of $41$ ant species measured at $30$ observational sites. In our experiment, we assumed responses are Poisson-distributed and we fitted GLLVMs using the \texttt{gllvm} package (top right), along with our proposed methods AIRWLS (bottom left) and Newton (bottom right). We found that all three methods capture qualitatively similar features of the observed response matrix (top left). This observation is confirmed in our quantitative analysis later on.}
    \label{fig:predictions}
\end{figure}

In order to validate our algorithms, we start by analyzing a small dataset of $41$ ant species measured at $30$ study sites in March--April 2008 \citep{gibb2011habitat}. Together with the abundance of ants at these sites, researchers measured a number of environmental variables at each site (\% cover of shrubs, bare ground, coarse woody debris, etc.). For a full report on the data acquisition methodology and the study design, \tred{we refer the reader to \citet{gibb2011habitat}. Here, we focus on illustrating how} GLLVMs can be used for identifying intrinsic environmental factors that are not expressed in measured habitat structure variables. 
\tred{For ordination purposes, we set the number of latent variables in the GLLVM to $p=2$. In Figure~\ref{fig:predictions}, we present the observed response matrix along with the predicted mean abundance for all species using the three methods of estimation.}

\begin{figure}[ht!]
    \centering
    \includegraphics[width=0.59\linewidth]{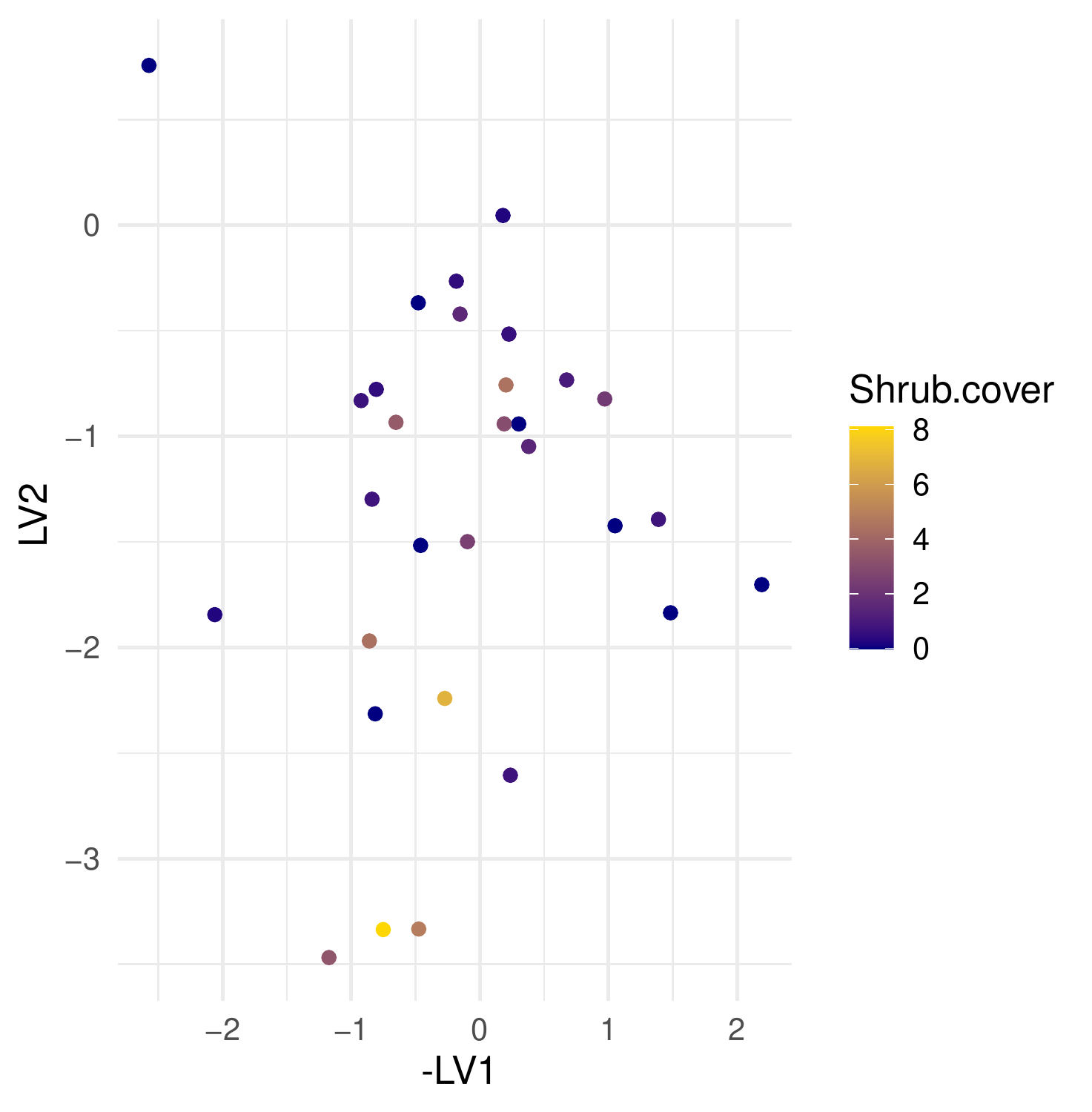}
    \includegraphics[width=0.4\linewidth]{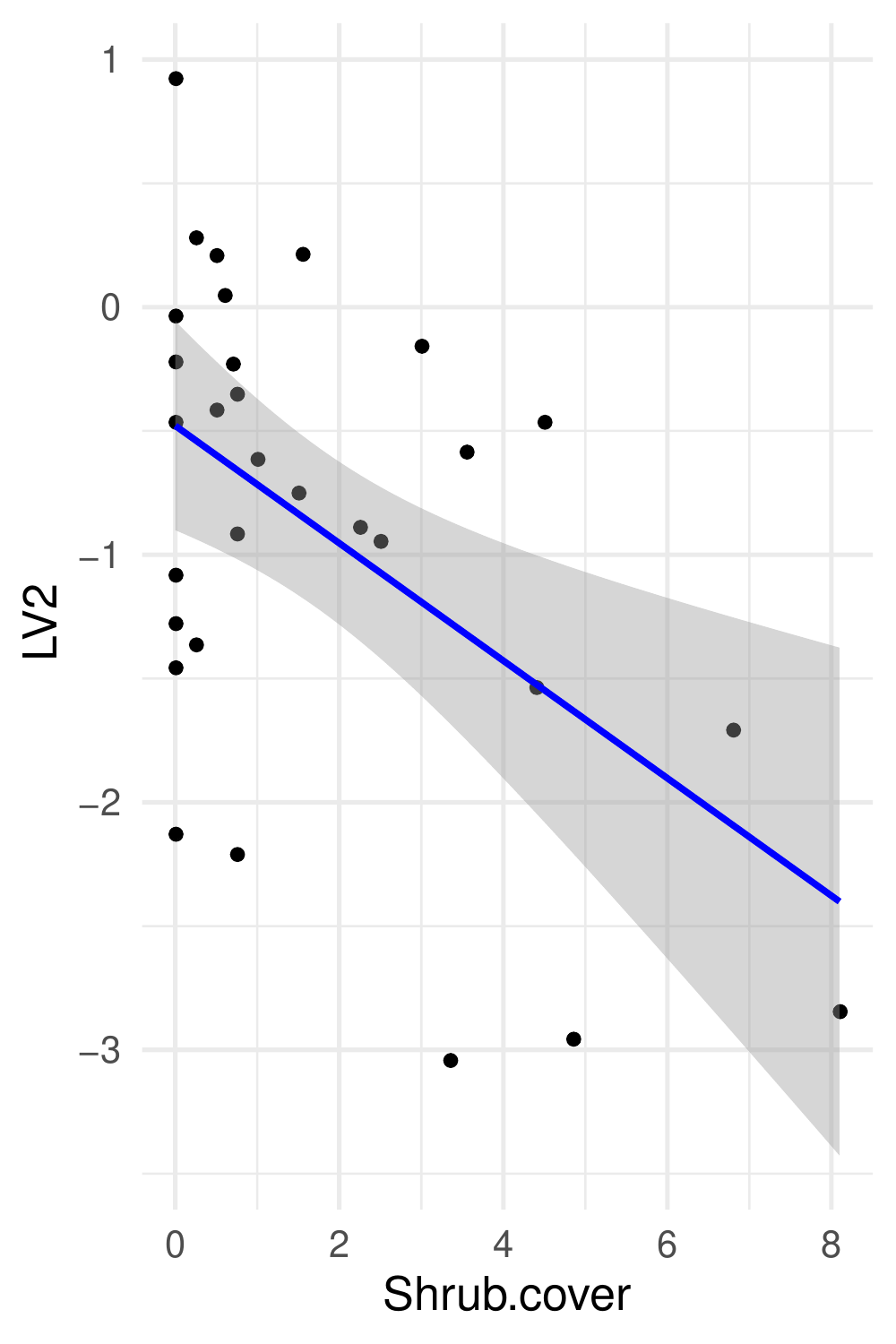}
    \caption{Validation of the generalized matrix factorization. In the dataset of abundance of ants, we held out a known feature of observational sites (\texttt{Shrub.cover}) and compared it to predicted latent variables/factor scores. We used the AIRWLS algorithm for model fitting. An ordination plot showing the latent space spanned by the estimated factor scores, colored by their corresponding value of \texttt{Shrub.cover}, is on the left panel. In the right panel, we found that the second predicted latent variable correlates strongly with the held-out predictor. }
    \label{fig:pc-plane}
\end{figure}

Next, to further validate the method, we removed one of the environmental variables and examined if some of its variability was captured by a latent variable (which could be interpreted as a missing covariate). That is, we hypothesized that some of variability explained by the variable can be explained by the inclusion of a latent variable. Here, we arbitrarily chose to remove \verb|Shrub.cover|. In Figure~\ref{fig:pc-plane}, we present an ordination plot of the predicted latent variable scores, as well as a scatterplot of the second latent variable versus \verb|Shrub.cover|. In the latter, the Pearson correlation coefficient between the two was $-0.49$, and it was statistically significant with $p$-value equal $0.005$ as based on a two-sided $t$-test. 

All three estimation approaches produced similar results in terms of deviance explained: \verb|gllvm| $79\%$, AIRWLS $79\%$, and Newton $75\%$. Turning to computation time the variational approximation approach using the \verb|gllvm| package fit converged in 2.3 seconds, while the quasi-Newton method took 0.2 seconds, and the AIRWLS algorithm converged in 0.5 seconds on 1 CPU and in 0.04 seconds when run in parallel across 8 CPUs. 

\subsection{Study 2: Large scale coexistence of species}\label{ss:species}

In this study, we analyzed data from systematic flora surveys along the east coast of New South Wales, using data obtained from the New South Wales Government (NSW Department of Planning Industry and Environment). Transects of the fixed area were exhaustively searched, and all plant species found in them were identified to species, where possible. We were interested in understanding co-occurrence patterns of different plants. A total of \num[group-separator={,}]{48737} transects were surveyed over the last two decades, and \num[group-separator={,}]{4841} species have been recorded from each transect. \tred{We first filtered out columns and rows with less than $0.1\%$ positive responses, leaving us with \num[group-separator={,}]{48331} observational units and 2,211 species. Each element in the response matrix is a binary response, reflecting the recorded presence-absence of a species at a specific location. Along with the species records, we also have $9$ covariates describing the local environment.}

\begin{figure}[ht!]
    \centering
    \includegraphics[height=3in]{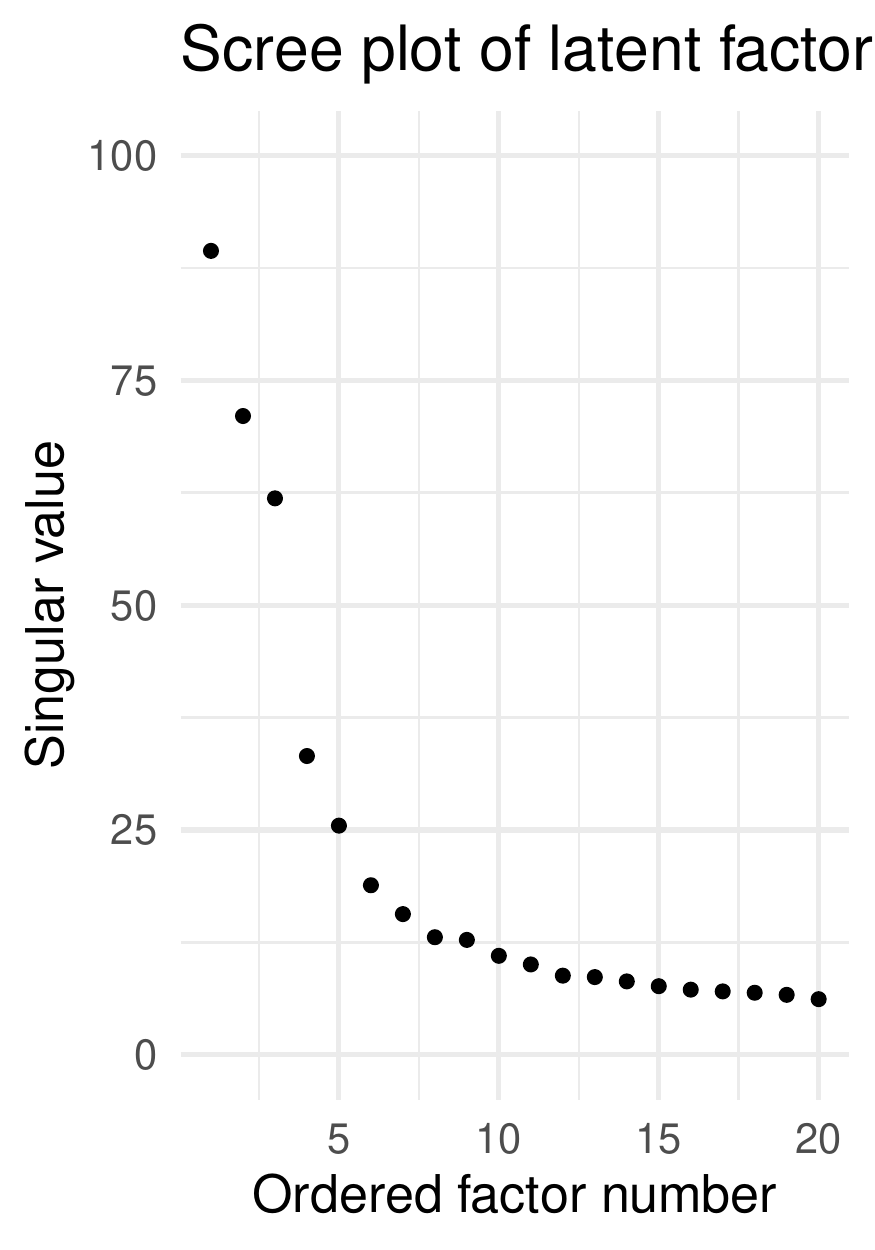}
    \includegraphics[height=3in]{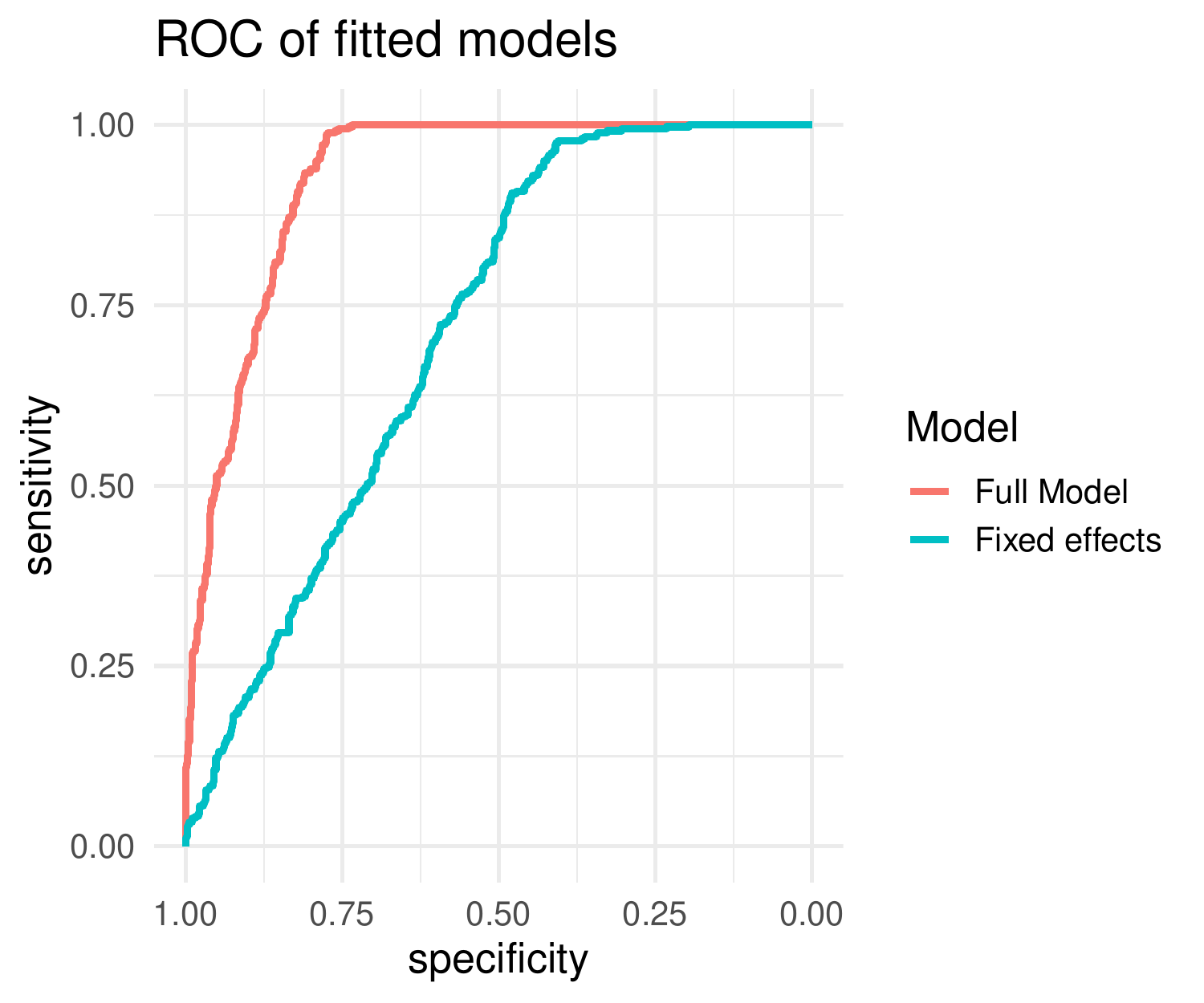}
    \caption{Model selection in the large-scale coexistence of species dataset. We used the quasi-Newton method to fit the model. To choose the number of factors, we used a scree plot, i.e., we plotted the singular values of the latent decomposition (left), defined as \tred{the squared roots of the diagonal elements of $\Lambda\Lambda^\intercal$}. The drop between the third and fourth values motivated the choice of the dimension of the latent space $p=3$. To validate if the latent space contained meaningful information, we compared ROC curves computed on a held-out dataset of 1,000 matrix entries and found a substantial increase in predictive power for the full model compared to the model only using fixed effects (right). }
    \label{fig:roc}
\end{figure}

\tred{First, to estimate the number of latent variables $p$, we} fitted a model with $p=20$ on the full dataset using the proposed quasi-Newton method in Section~\ref{ss:quasi-newton}\tred{, and looked for a drop-off in size of the singular values from this fit.} Based on the scree plot (Figure~\ref{fig:roc}, left panel) we set $p=3$.

\tred{Next we evaluated the predictive performance of estimated model (in particular, to assess the importance of including latent variables) by leaving out 1,000 elements in the response matrix and computing out-of-sample measures of predictive performance, as described in Section~\ref{ss:missing-data}. Specifically, we} held out $500$ elements for which the model predicted $1$ with at least $0.5$ probability and $500$ elements for which the model predicted $0$ with at least $0.5$ probability. We evaluated the fit by calculating the out-sample deviance and area under the receiver operator curve (AUC), and \tred{compared performance from only the fixed effect model i.e., with $p = 0$, and the full model with $p = 3$ latent variables (Figure~\ref{fig:roc}, right panel)}.

Due to the large scale of the problem, we were unable to use the \verb|gllvm| package. Therefore, we tested only the Newton algorithm and it \tred{converged} in three hours on commodity hardware. Our model with only fixed effects had AUC = $0.72$ and explained $39\%$ of the out-of-sample deviance in the held-out dataset. By contrast, the full model containing $p = 3$ latent variables models had AUC = $0.87$ and explained $58\%$ of the out-of-sample deviance (Figure~\ref{fig:roc}, right panel).

\tred{Next, we examined how the size of the dataset influenced the performance of the various estimation algorithms. We sub-sampled rows and columns from the response matrix and fitted models to the subsetted data. Specifically, we set} $\rho \in \{ 0.005, 0.01, 0.015, ..., 0.065 \}$ and sampled $\left\lfloor\rho\cdot n\right\rfloor$ rows and $\left\lfloor\rho\cdot m\right\rfloor$ columns, i.e. from $0.5\%$ to $6.5\%$ of the total number of rows and columns in $Y$. \dw{We did not use more than $6.5\%$ of rows and columns because at this size the \texttt{gllvm} package was already having difficulty converging to a solution.} Based on the scree plot of singular values of the full model (Figure~\ref{fig:roc}), we fixed the number of latent variables at $p=3$ across this simulation. Also, in order to use a metric comparable across different samplings, we used the mean deviance i.e., $D(Y,\hat M)$ standardized by the number of rows and columns in the corresponding response matrix.

\begin{figure}[ht!]
    \centering
    \includegraphics[width=0.37\linewidth]{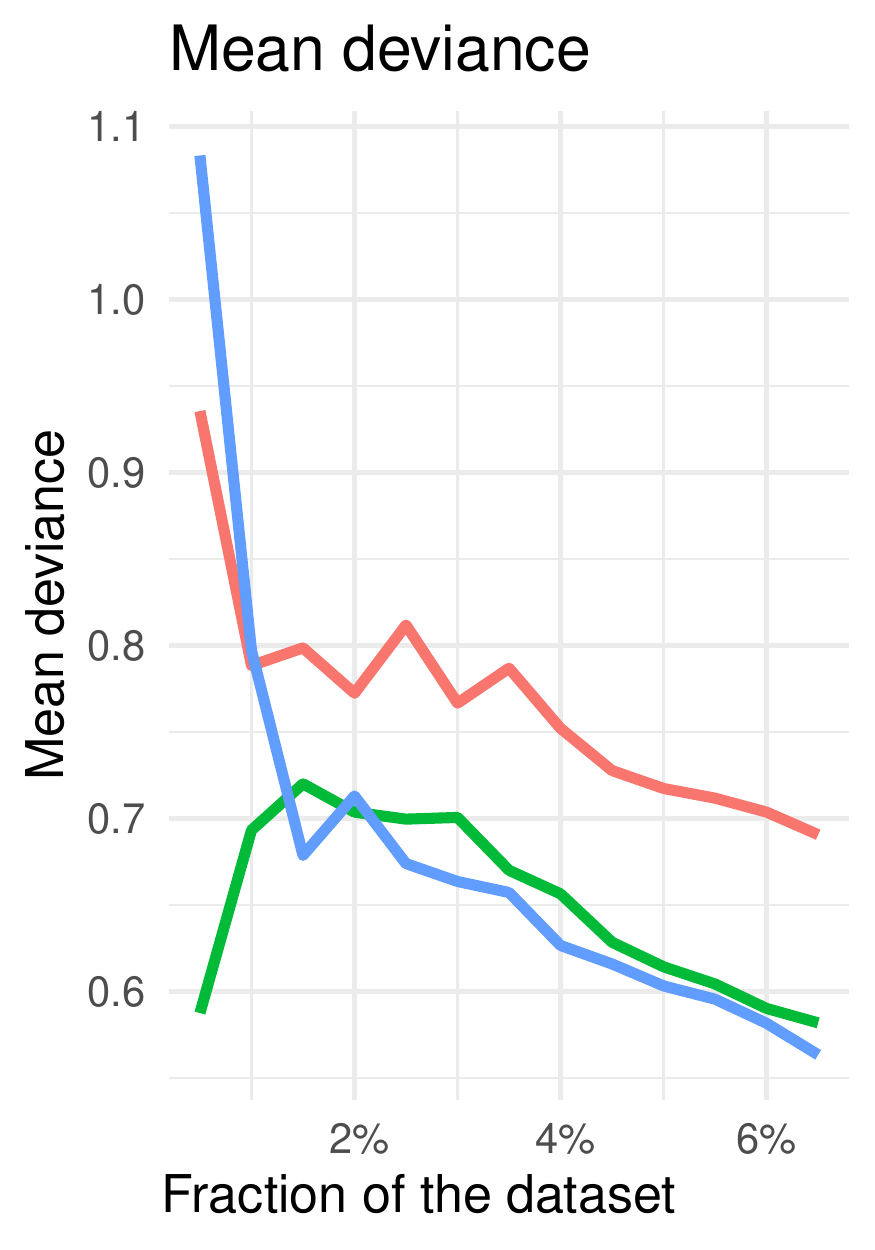}
    \includegraphics[width=0.58\linewidth]{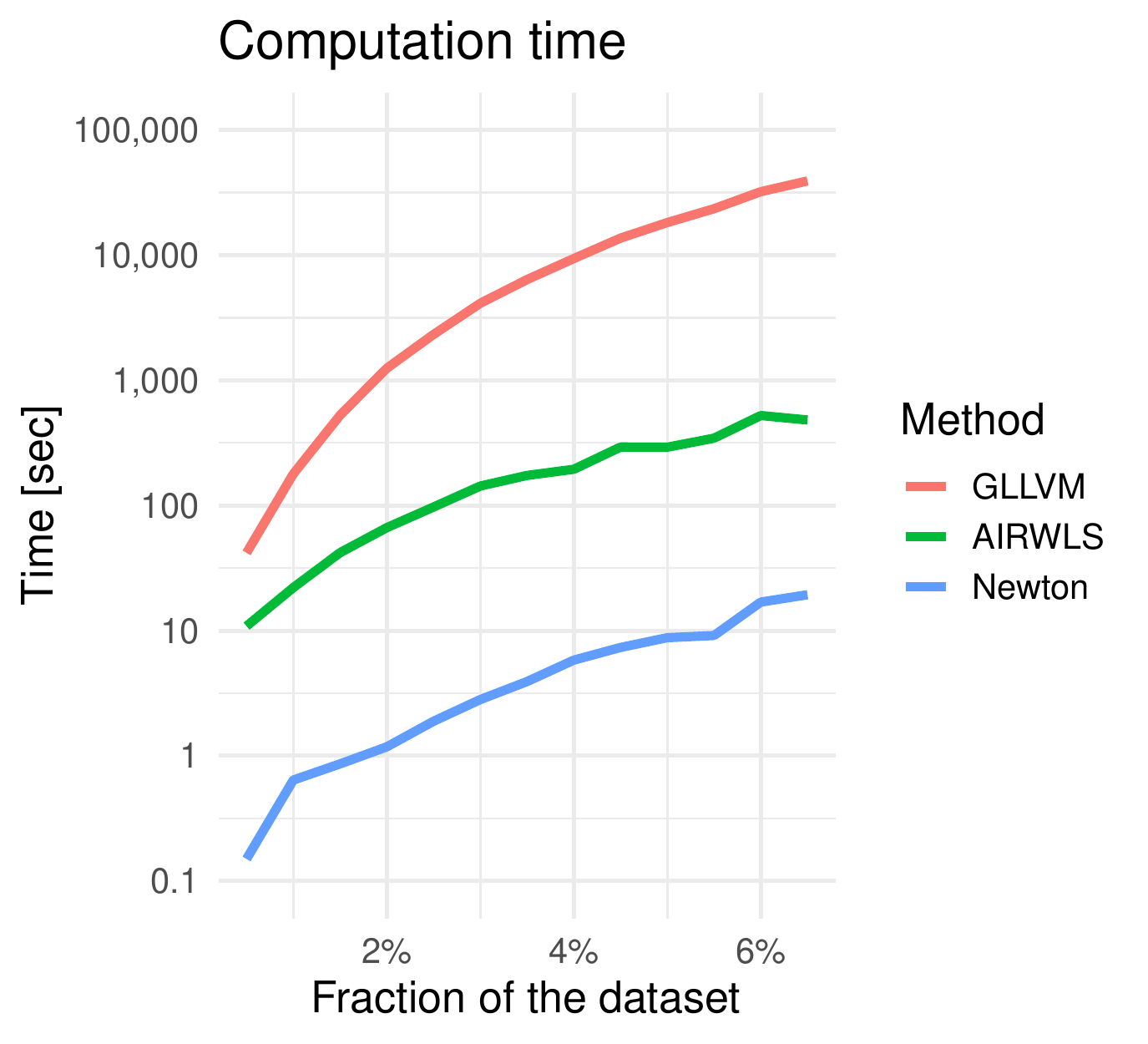}
    \caption{Mean deviance and computation time as a function of the dataset size. We sampled fractions of columns and rows from the large dataset of coexistence of species, and compared three methods for fitting GLLVMs to the sub-sampled subsets. A fraction $\rho \in \{0.01,...,0.065\}$ corresponded to a response matrix with $\left \lfloor\rho\cdot 48737\right\rfloor$ rows and $\left\lfloor\rho\cdot 2211\right\rfloor$ columns. We fixed the number of latent variables at $p=3$. Both AIRWLS and Newton methods outperformed the baseline \texttt{gllvm} implementation both in terms of explained deviance (left) and computation time (right). \tred{We only report results up to 6.5\% of the dataset, since applying the baseline \texttt{gllvm} to larger datasets was not computationally feasible. We note however that the quasi-Newton method converged in three hours on the full dataset.}}
    \label{fig:row-col-sampling}
\end{figure}

\tred{The proposed quasi-Newton method performed best both in terms of deviance and computation time, although for larger fractions of the full dataset AIRWLS performed very similarly in terms of mean deviance} (Figure~\ref{fig:row-col-sampling}). Elementary linear extrapolation suggests that computing a solution for the full dataset via \verb|gllvm| would take at least 2 weeks \dw{(if it converged at all)}. In light of this, we chose \dw{not to attempt to fit} \verb|gllvm| to the full dataset due to anticipated memory constraints.  

\section{Simulations}\label{s:simulations}

We conducted a numerical study to empirically investigate how the number of responses, the number of latent variables, and the distribution of responses influence the performance of the proposed algorithms for estimating GLLVMs.

\subsection{Simulation design}\label{ss:sim-setting}

We designed our simulations to mimic the setting from the data study described in Section~\ref{ss:ants}. To that end, we started with a GLLVM fitted to the ant abundance data. We used sample estimates of the covariance matrix of $X$ and $\Lambda$ as the basis for sampling new multivariate Gaussian variables for $X$ and $\Lambda$. For $U$ and $B$, we constructed matrices by randomly simulating each element independently from a standard Gaussian distribution. \tred{Using these quantities, we were able to construct a matrix of fitted values $\mu_{ij}$, from which we can then simulate a response matrix following \eqref{eq:model}. We considered different combinations of the number of observational units $n$, the number of responses in each unit $m$, the number of latent variables $p$, and the distribution of responses (assuming a canonical link all response distributions). In particular, we ran simulations with $n,m \in \{100, 200, 300, 500\}$, $p \in \{2, 3, 5, 10\}$, and considered Poisson and binomial distributions. Additionally, we tested a small sample setting with $n\in\{5,10,15\}$, $m=20$, and $p=2$. For each set of simulation parameters, we repeated the experiment with $100$ generated datasets.}



As in Section~\ref{s:data}, for each simulated dataset, we fitted the GLLVM using the \verb|gllvm| package, along with the proposed quasi-Newton and AIRWLS methods. For all three methods, we used the same stopping criterion, with a relevant error tolerance equal to $10^{-3}$. We evaluated performance using mean deviance, Procrustes error for the fit of the latent space, and mean squared error of the estimated response-specific coefficients, all of which were detailed in Section~\ref{ss:evaluation}.

\subsection{Results}
\begin{figure}[ht!]
    \centering
    \includegraphics[width=0.35\linewidth]{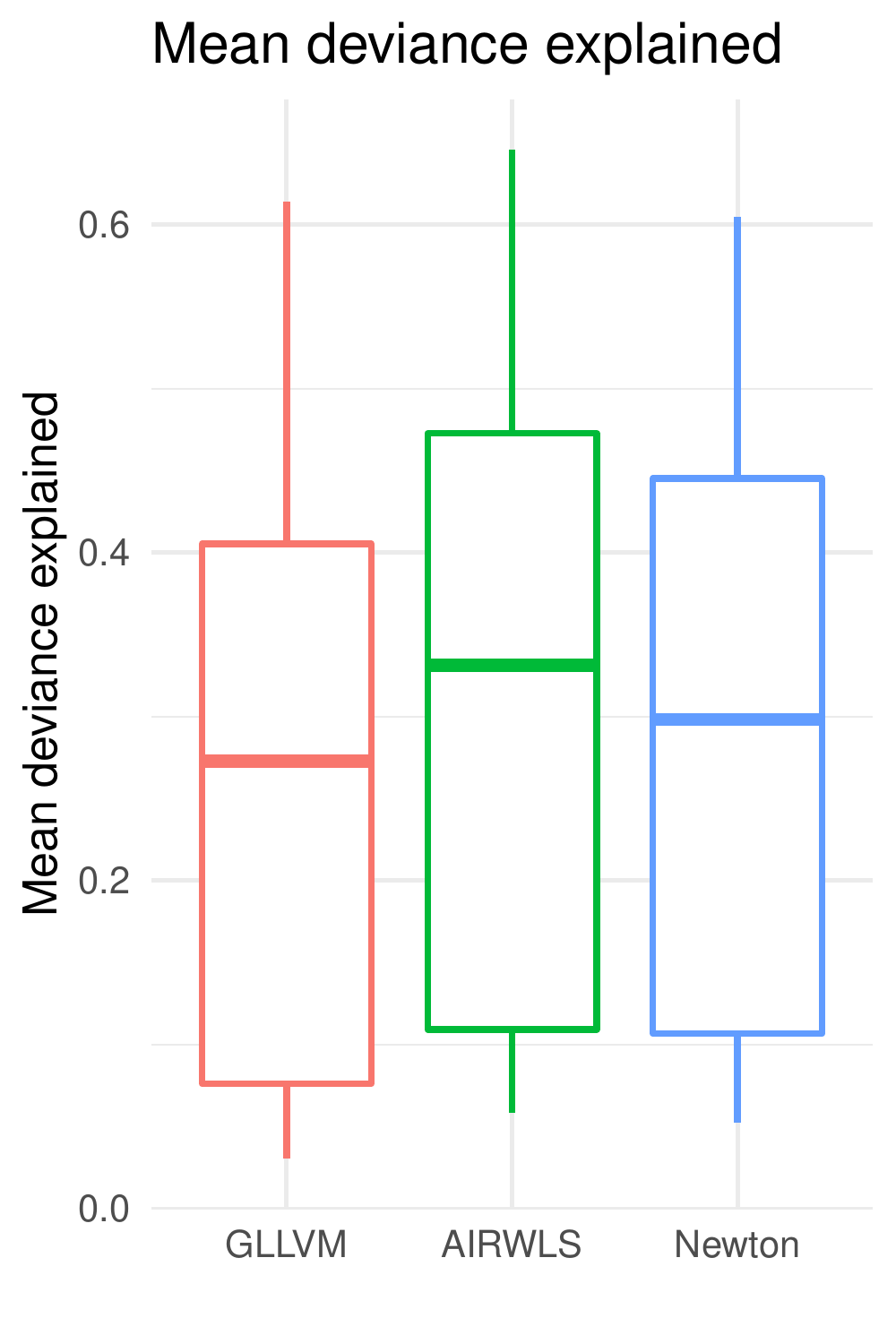}
    \includegraphics[width=0.35\linewidth]{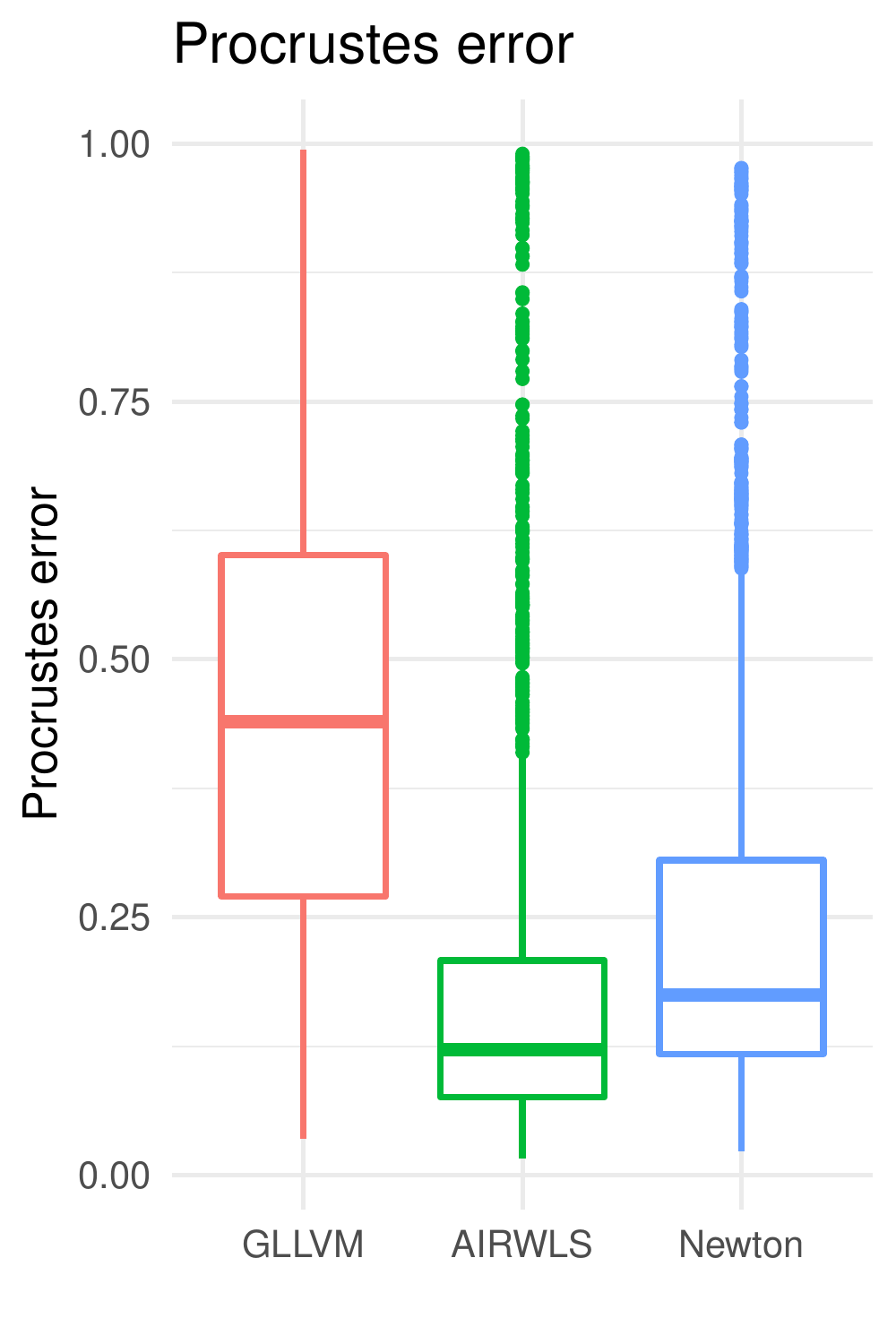}\\
    \includegraphics[width=0.35\linewidth]{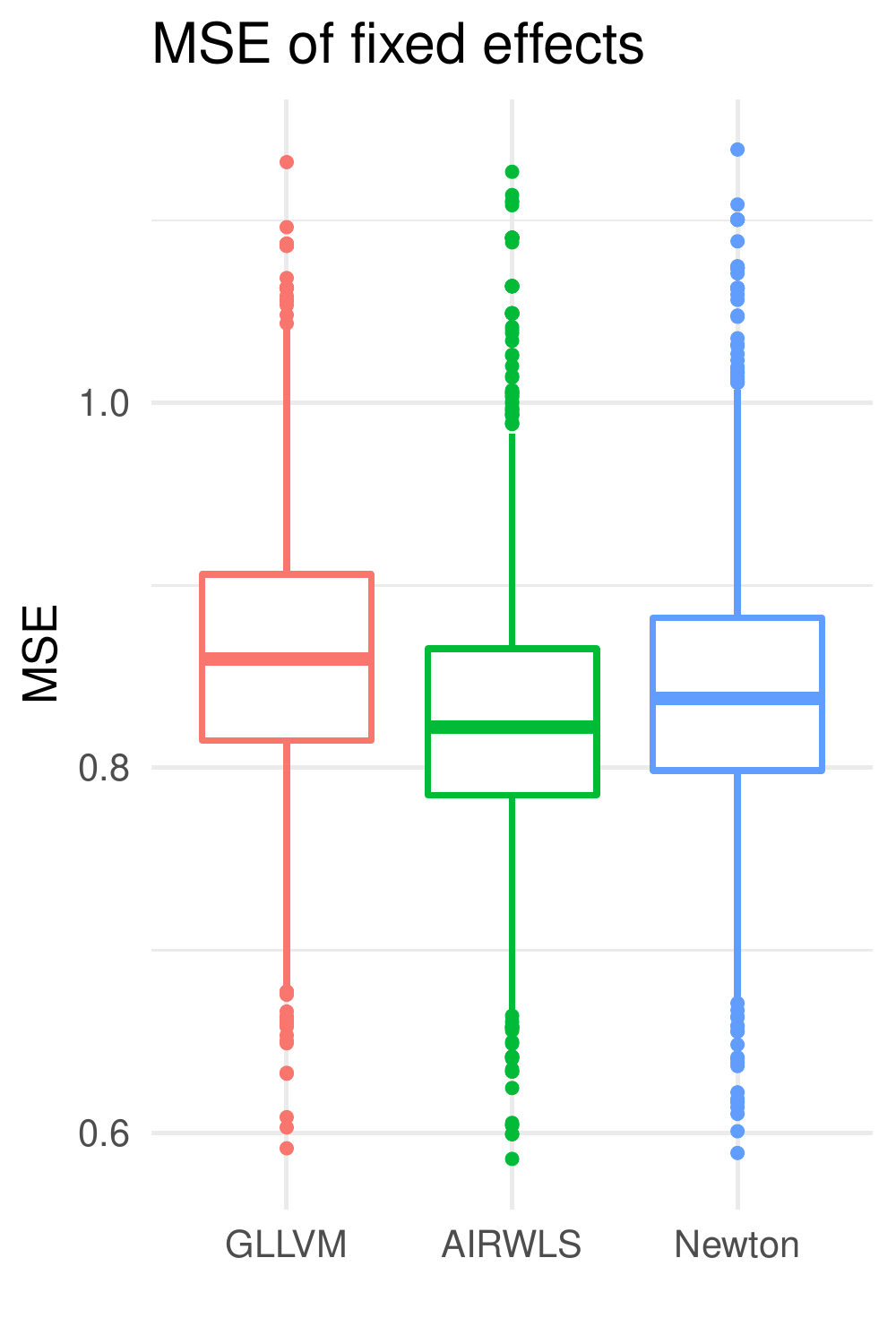}
    \includegraphics[width=0.35\linewidth]{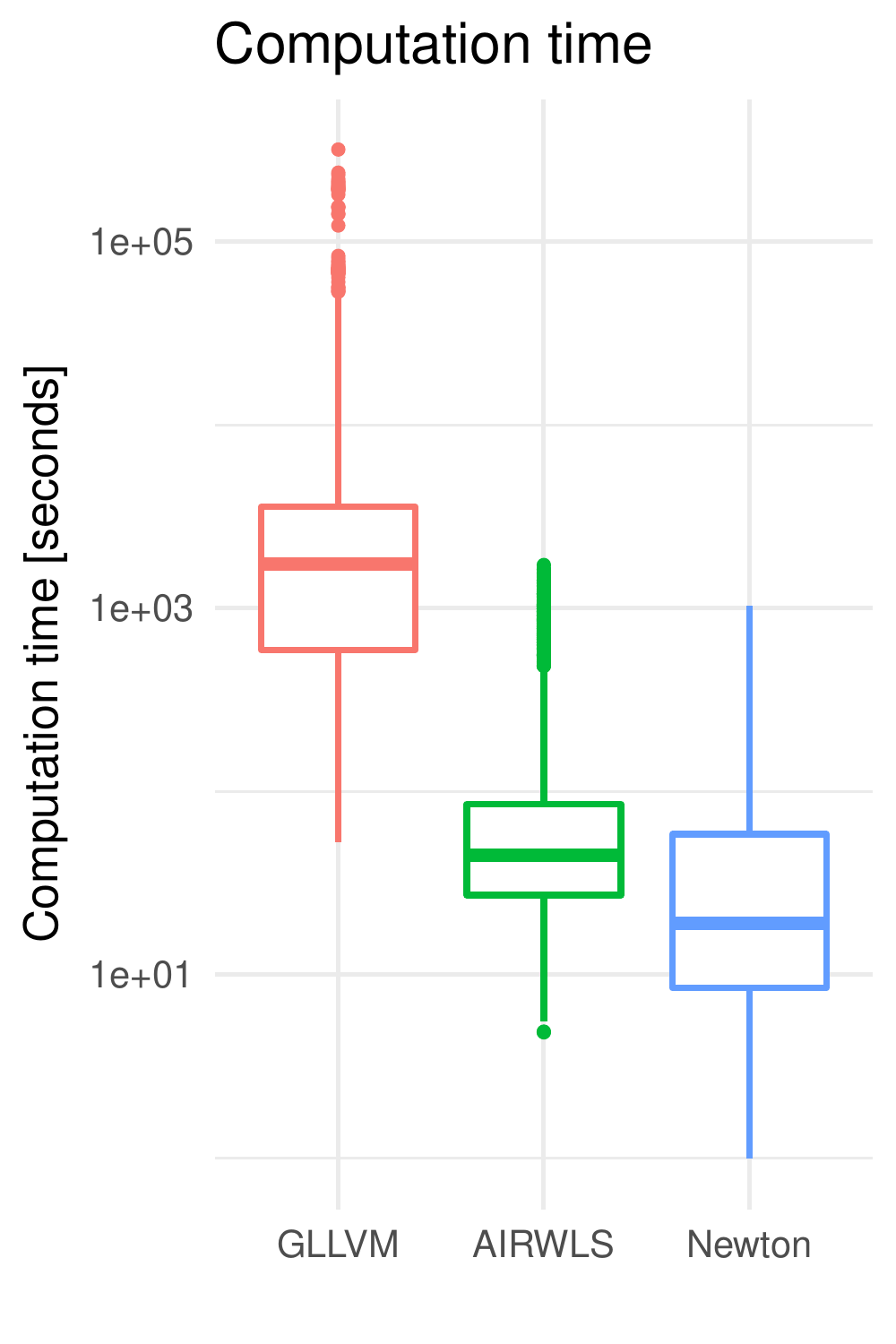}\\
    \caption{Comparison of the proposed AIRWLS and quasi-Newton methods with the \texttt{gllvm} package for fitting GLLVMs. In this simulation study, we sampled data such that covariances of $X$ and $V$ matched those of the sample estimates from the ant abundance data. \tred{We varied $n,m \in \{100, 200, 300, 500\}$, $p \in \{2,3,5,10\}$ and the distribution of responses (Poisson or Binomial)}. Based on generating 100 datasets for each combination of simulation parameters, we found that the AIRWLS slightly outperformed other methods on all metrics (mean deviance explained, Procrustes error, and MSE of fixed effects), while the Newton algorithm \dw{was superior in terms of computation} time.} \label{fig:sim-results}
\end{figure}

Aggregating across the combinations of $n, m$ and $p$, considered, both \verb|gllvm| and the quasi-Newton method produced similar performance in terms of deviance explained, Procrustes error, and the MSE of fixed effects, while AIRWLS performed slightly better across these three measures (Figure~\ref{fig:sim-results}). The main gain of the proposed methods, however, comes in computation time: using the same stopping criteria, computation times differed by several orders of magnitude, and on average the quasi-Newton algorithm took $51$ seconds to compute, AIRWLS method took $132$ seconds, while \verb|gllvm| $51.6$ minutes (Figure~\ref{fig:sim-results}, bottom right).

\begin{figure}[ht!]
      \centering
      \includegraphics[width=0.8\linewidth]{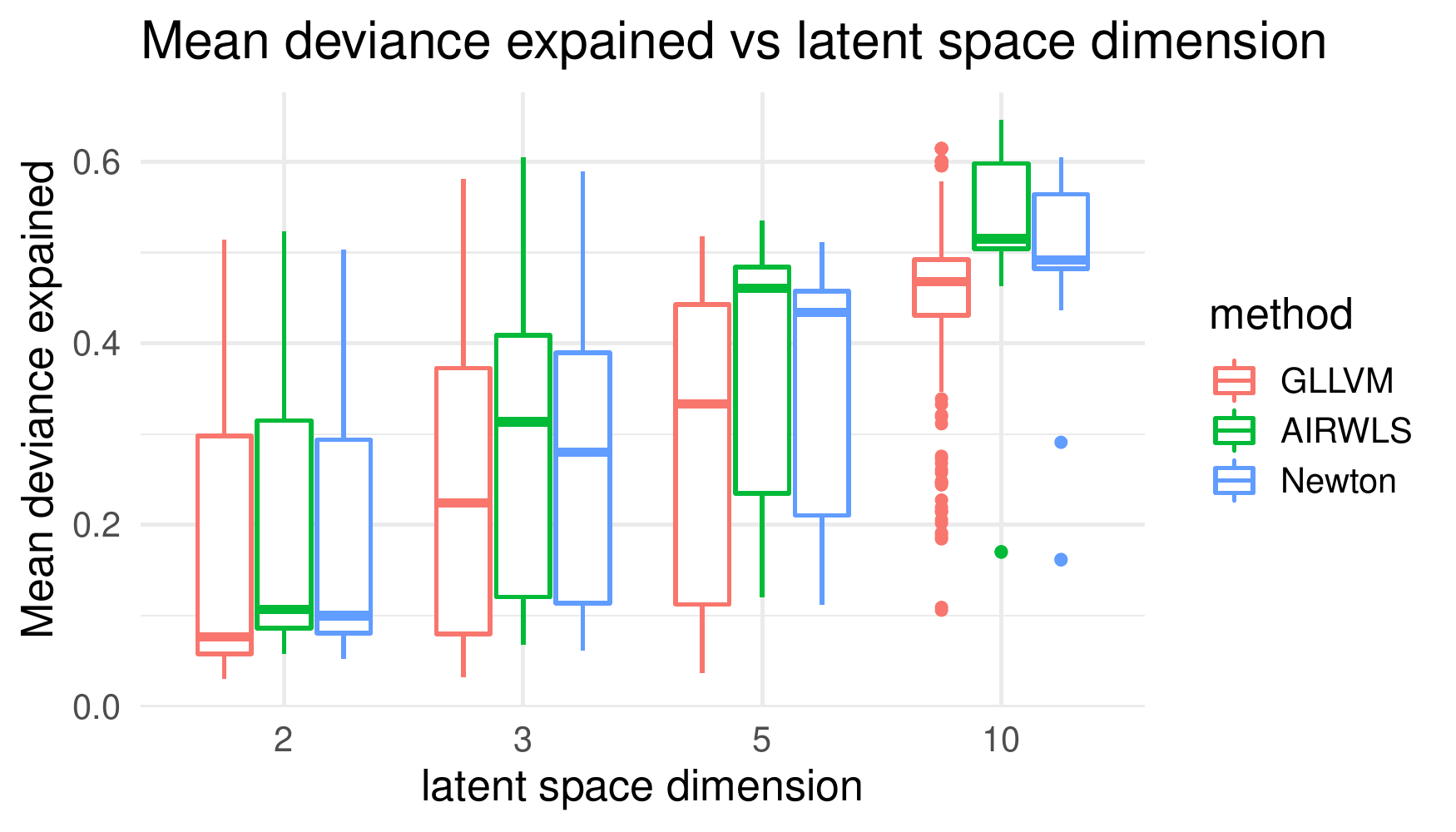}
      \includegraphics[width=0.8\linewidth]{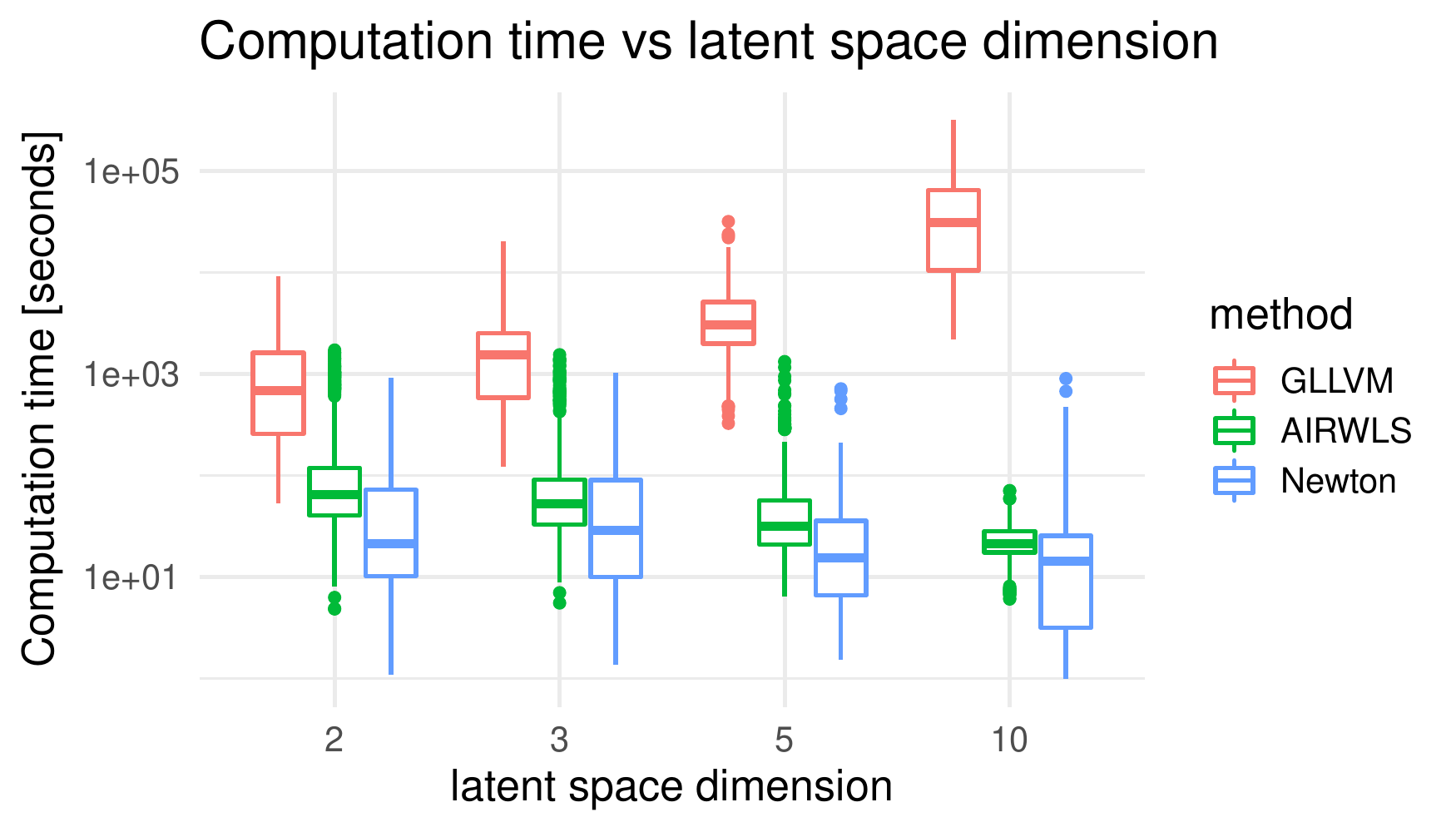}
      \caption{\tred{Assessment of how the number of latent variables $p$ affects computation time and accuracy in terms of deviance explained. We found that as $p$ grows, the proposed methods methods became more accurate \dw{than \texttt{gllvm}}, while the difference in computation time also grew.}}
      \label{fig:results-p}
    \end{figure}

\tred{Next, we investigated how the size of the response matrix, number of latent variables, and response distribution affected model performance and computation time. Overall, we found that in the small data experiment ($m=20, n\in\{5,10,15\}$), variational approximations using the \texttt{gllvm} packaged outperformed the proposed AIRWLS and quasi-Newton methods for $n=5$, with deviance explained equal to $0.922$, $0.733$ and $0.65$ respectively. However, all methods performed similarly in terms of mean deviance \dw{explained once} $n > 10$.} \tred{In terms of the number of latent variables, we found that as $p$ increased both AIRWLS and quasi-Newton methods started to outperform the \texttt{gllvm} package in terms of deviance explained, and the gap in computation time increased in their favor (see Figure \ref{fig:results-p}). An extreme example of this is at $p=10$, $n=500$, and $m=300$, where we found that the \texttt{gllvm} package took approximately 2.5 days to converge, while AIRWLS converged in 40 seconds and resulted in a more accurate fit in terms of deviance explained.}

\tred{Finally, GLLVMs with binary responses (and using the logistic link) tended to be more difficult to fit \dw{than the} Poisson distributions, and this occurred for all three estimation methods we \dw{tested}. For example, keeping all other model parameters the same, the average mean deviance explained for GLLVMs fitted using \texttt{gllvm}, AIRWLS, and Newton was 0.383, 0.434, and 0.412 respectively for \dw{the} Poisson distribution, but this decreased to 0.092, 0.127, and 0.125 respectively for \dw{the} binomial distribution. The generally poorer fit observed for binary responses is consistent with what is commonly observed for \dw{GLMs, and} reflects the general lack of information in binary response as well as potential issues such as quasi or complete-separation in the fitting process.}

\FloatBarrier

\subsection{Model selection: Unknown $p$ and regularization}\label{ss:unknown-d}

Our cross-validation framework introduced in Section~\ref{ss:evaluation} allows us to choose the tuning parameters, particularly for predictive applications. As an illustration of this, we simulated responses using the process described in Section~\ref{ss:sim-setting}, \tred{and compared two approaches for constraining and choosing the dimension of the latent space: smooth regularization based on sliding the parameter $\gamma$ as described in Section~\ref{ss:reg-gmf}, and an explicit rank constraint based on selecting $p$ from a set of candidate integer values. We \dw{used 20-fold cross-validation to select both $\gamma$ and} $p$, and used out-of-sample deviance as the performance measure. Finally, we compared the predictive performance based on these selected parameter values by fitting the corresponding GLLVMs to a full training set containing 95\% of the observations and then computing the deviance on the remaining hold-out set containing 5\% of the observations.}

\tred{In more detail, we set $n=m=100$, the true dimension of the latent space as $p=2$, and simulated Poisson responses. In both model selection approaches, we used the quasi-Newton method for fitting GLLVMs, and kept all parameters other than $p$ and $\gamma$ equal. We implemented smooth regularization as defined in \eqref{eq:two-penalties} with $\gamma \in \{0,1,...,60\}$, while for the rank-constrained models we considered $p \in \{1,2,...,50\}$.}

\tred{Based on repeating the entire simulation above twenty times, we found that both methods of model selection achieved very similar results, with mean deviance $1.272$ ($sd=0.130$) for the smooth regularization \dw{approach}, and $1.279$ ($sd=0.187$) for the rank-constrained approach.} We conclude (in particular) that the cross-validation and smooth regularization approach for selecting $p$ is a promising approach for tuning GLLVMs, made possible by the computational gains from our PQL-based estimation methods. We leave the thorough analysis of the theoretical properties of cross-validation and regularization of $\gamma$ as an avenue of future research.


\section{Discussion}
For estimating GLLVMs, our proposed PQL-based methods are orders of magnitude faster than current state-of-the-art algorithms for estimating model parameters for GLLVMs. \tred{They can be decomposed and parallelized across multiple \dw{machines}, leveraging modern open parallel computing platforms such as Apache Spark \citep{zaharia2010spark}.} \tred{It should be emphasized though that PQL-based methods are expected to work well in the settings \dw{that motivated} this article, i.e., fitting GLLVMs to high-dimensional datasets. While estimates from PQL can exhibit more finite sample bias relative to say, the Laplace \dw{method}, it has nevertheless been proven that they generally perform well when the number of \dw{observations per random effect, $m$ in our setting}, becomes large \citep{nie2007convergence,hui2017joint,hui2021use}. This is precisely the \dw{scenario of interest here}, since one of the main reasons behind fitting a latent variable model, in general, is to perform matrix factorization and approximate a covariance matrix between $m$ responses when $m$ is too large to use an unstructured estimate. 
}

Our algorithms are elementary to implement using existing GLM routines. Specifically, we provide an \verb|R| implementation via the open-source package \verb|gmf|\footnote{\url{http://github.com/kidzik/gmf/}} enabling integration with existing workflows and further extension of our algorithms.  \tred{Our package can be used as a drop-in replacement for widely-adopted heuristic methods such as, for example, fitting principal component analysis to log-transformed count data. }

Throughout this work, we have illustrated the applicability of our methodology in the context of ecology. However, similar problems can be found in other disciplines whenever we are interested in extracting latent factors underlying certain responses. Our methods are particularly useful when matrices are large, for example, in studies of behavior of subjects online with thousands of individuals and items or web pages they view, \tred{or count data in single-cell RNA sequencing.}

\section{Acknowledgements}
Łukasz Kidziński was supported by the Mobilize Center grant U54 EB020405 from the National Institute of Health. Francis K.C. Hui was supported by an Australian Research Council Fellowship (DE200100435). David I. Warton was supported by the Australian Research Council’s Discovery Project Scheme (project DP210101923). Trevor J. Hastie was partially supported by grants DMS-2013736 And IIS, 1837931 from the National Science Foundation, and grant 5R01 EB, 001988-21 from the National Institutes of Health.

\bibliography{references}



\end{document}